%% file: main.tex
\documentclass{article}

\usepackage[final]{corl_2026} 

\usepackage{amsmath} 
\usepackage{algorithm}
\usepackage{algpseudocode}
\usepackage{graphicx}
 \usepackage{booktabs}

\usepackage{hyperref}
\usepackage{url}
\usepackage{footmisc}

\usepackage{amsmath}
\usepackage{amssymb}
\usepackage{mathtools}
\usepackage{amsthm}
\usepackage{tabularx}
\usepackage{utfsym}

\definecolor{ForestGreen}{rgb}{0.1333,0.545,0.1333}
\definecolor{Firebrick}{rgb}{0.698,0.1333,0.1333}
\usepackage{caption}

\usepackage{subcaption}
\usepackage{caption}
\usepackage{placeins}
\usepackage{xspace}
\usepackage{multirow}     
\usepackage[dvipsnames]{xcolor}

\definecolor{ForestGreen}{rgb}{0.1333,0.545,0.1333}
\definecolor{Firebrick}{rgb}{0.698,0.1333,0.1333}

\usepackage{xcolor}  

\definecolor{robotaction}{RGB}{255, 140, 0}  
\definecolor{robottype}{RGB}{178, 34, 34}
\definecolor{robottask}{RGB}{65, 105, 225}  
\definecolor{controltype}{RGB}{0. 205, 0}
\definecolor{predsteps}{RGB}{216, 191, 216}
\definecolor{lavendermist}{rgb}{0.9, 0.9, 0.98}
\definecolor{visualtrace}{RGB}{255, 216, 0} 
\usepackage{colortbl}
\usepackage{xcolor}
\definecolor{lightgray}{gray}{0.9}
\usepackage{booktabs}
\usepackage{color, soul}
\usepackage{graphicx}
\usepackage{amsmath}
\usepackage{empheq}
\usepackage{cleveref}
\usepackage{epigraph}
\definecolor{lightgray}{gray}{0.9}
\definecolor{lightblue}{rgb}{0.93,0.95,1.0}
\definecolor{darkgreen}{rgb}{0.0,0.6,0.0}
\definecolor{darkblue}{rgb}{0.0,0.0,0.5}
\definecolor{pinegreen}{rgb}{0.0, 0.47, 0.44}
\definecolor{deepmagenta}{rgb}{0.8, 0.0, 0.8}
\definecolor{amber}{rgb}{1.0, 0.49, 0.0}

\definecolor{checkgreen}{RGB}{34,139,34}
\definecolor{crossred}{RGB}{200,40,40}

\newcommand{\cmark}{\textcolor{checkgreen}{\ding{51}}}
\newcommand{\xmark}{\textcolor{crossred}{\ding{55}}}

\newcommand{\ignorebig}[1]{}

\newcommand{\minisection}[1]{\noindent{\textbf{#1}.}}

\newlength\savewidth

\definecolor{lightred}{RGB}{241,140,142}
\definecolor{amber(sae/ece)}{rgb}{1.0, 0.49, 0.0}
\definecolor{battleshipgrey}{rgb}{0.52, 0.52, 0.51}
\definecolor{cadmiumorange}{rgb}{0.93, 0.53, 0.18}
\definecolor{applegreen}{rgb}{0.55, 0.71, 0.0}
\definecolor{cadmiumgreen}{rgb}{0.0, 0.42, 0.24}
\definecolor{forestgreen}{rgb}{0.13, 0.55, 0.13}
\definecolor{red}{rgb}{0.89, 0.0, 0.13}

\definecolor{cb-0}{RGB}{216, 27, 96}
\definecolor{cb-1}{RGB}{30,136,229}
\definecolor{cb-2}{RGB}{255,193,7}
\definecolor{cb-3}{RGB}{0, 77, 64}
\definecolor{cb-4}{RGB}{150,220,174}

\input{preamble}
\title{Social Humanoid: Socially Responsive Human–Humanoid Interaction}

\author{
\parbox{\textwidth}{\centering\normalfont
Vongani H. Maluleke$^{\dagger *}$ \quad James Ni$^{\dagger *}$ \quad Kie Horiuchi$^{\dagger \ddagger}$ \quad Lea Wilken$^{\S}$ \quad Jiahui Lei$^{\dagger}$ \\\quad {Evonne Ng}$^{\S}$ \quad Roei Herzig$^{\dagger}$ \quad Jitendra Malik$^{\dagger}$ \quad Trevor Darrell$^{\dagger}$  \quad Angjoo Kanazawa$^{\dagger}$ \\[0.6em]
UC Berkeley$^{\dagger}$, Sony Group Corporation$^{\ddagger}$, Meta$^{\S}$}
}

\begin{document}
\maketitle
\begingroup
\renewcommand{\thefootnote}{}
\footnotetext{*These authors contributed equally. Correspondence to: \texttt{vongani\_maluleke@berkeley.edu}}
\endgroup

\begin{figure*}[ht!]
    \vspace{-1cm}
    \centering
    \includegraphics[width=1\textwidth]{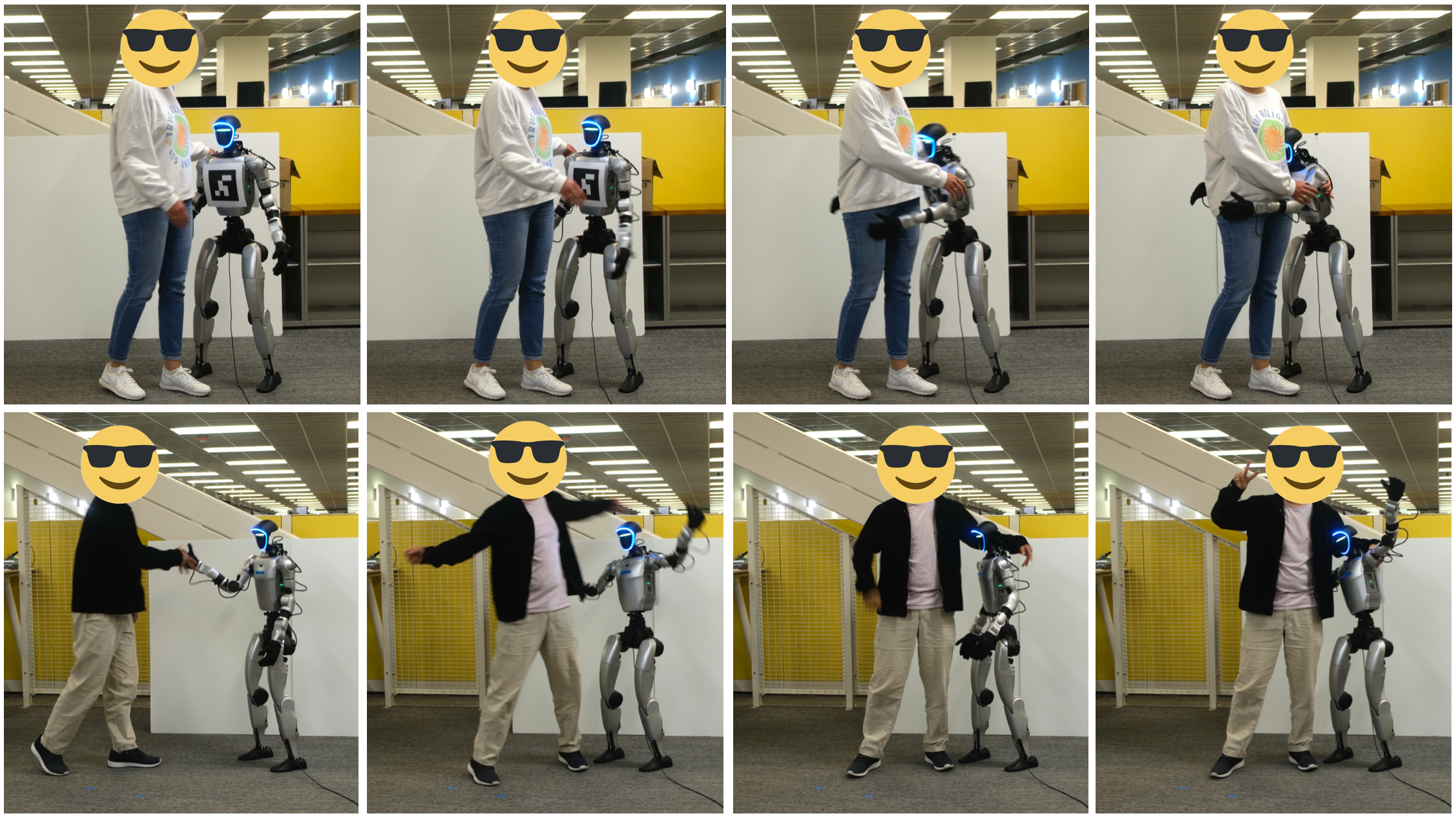}
    \caption{ \textbf{\ourmethod{}} enables humanoids to perceive, anticipate, and respond to people with coordinated whole-body behavior. Examples include a hug (top) and a greeting followed by posing together for a photo (bottom). Given RGB-D video, our method reconstructs the observed human motion, then conditions on it to predict future human motion and a humanoid response. The generated motions are executed by a low-level tracking controller. Video results are available on our webpage: \href{https://von31.github.io/social-humanoid/}{https://von31.github.io/social-humanoid/}
    }
    \vspace{-6px}
    \label{fig:teaser}
\end{figure*}

\input{sec/00_abstract}
\input{sec/01_intro}
\input{sec/02_related_work}
\input{sec/03_method}
\input{sec/04_experiments}
\input{sec/05_conclusion}

\clearpage

\acknowledgments{
We thank Zeldah Maluleke, Kaito Suzuki, Marsalis Gibson and Sharmita Dey for serving as the human partner/operator in our real-world demonstrations and assisting with recording the demonstrations, and Zekai Wang and our BAIR colleagues for helpful discussions. This project was funded in part by Sony Corporation, Bakar Fellows, NSF CAREER award no 2442491, the Humanoid Intelligence Center program, ONR MURI N00014-21-1-280, and MURI N000142612024.

}


\bibliography{main}  

\clearpage
\appendix
\input{sec/supp}

\end{document}

%% file: preamble.tex
\newcommand{\ourmethod}[1]{Social Humanoid}

  \usepackage{makecell}      
  \usepackage{amssymb}       
  \usepackage{pifont}        
  \usepackage[table]{xcolor}
  \usepackage{array}         
\usepackage{multirow}
\usepackage{tabularx}
\usepackage{colortbl} 
\usepackage{caption}
\usepackage{wrapfig}
\usepackage{minitoc}

\newcolumntype{C}{>{\centering\arraybackslash}X}

\newcommand{\best}[1]{\ifmmode\bm{#1}\else\textbf{#1}\fi} 

\makeatletter
\newcommand{\fullrowcolor}[1]{\noalign{%
  {\color{#1}\hrule height \ht\@arstrutbox depth \dp\@arstrutbox}%
  \vskip-\dimexpr\ht\@arstrutbox+\dp\@arstrutbox\relax}}
\makeatother

%% file: sec/00_abstract.tex
\begin{abstract}

Humanoid robots are typically trained as solitary agents, leaving them unable to coordinate physically with human partners who naturally anticipate and adjust to one another.
To bridge this gap, we present \ourmethod{}, a hierarchical system that generates and executes socially responsive whole-body motion on a 29-DoF Unitree G1 conditioned on a human partner's motion.
Its high-level policy performs flexible sequence modeling of human and humanoid motion through a shared tokenized representation, generating reactive humanoid motion and anticipating the partner's motion when they are occluded, while a low-level whole-body tracking controller executes the generated motion on the robot.
To train the system, we introduce a two-phase adaptive retargeting procedure that retargets one interaction partner to the robot while morphing the other partner to preserve contact dynamics, reachability, and relative body scale.
We apply this procedure across five motion-capture datasets to produce a new Human-Humanoid Interaction (HHI) benchmark of 260 hours of human-humanoid interactions.
Our results demonstrate that \ourmethod{} produces coordinated, human-aligned behavior across diverse interaction types in simulation and the real world. 
\end{abstract}

\keywords{Human-Humanoid Interaction, Motion Retargeting}

%% file: sec/01_intro.tex
\section{Introduction}
Human interaction is the art of attunement, a continuous interplay of perception, anticipation, and response that enables seamless coordination. Whether synchronizing a high-five or collaborating on a shared task, people adjust effortlessly through gaze, posture, and timing, without words. This ability is a hallmark of social intelligence~\cite{thorndike1920intelligence, gardner1983frames, goleman2006social}, allowing humans to simultaneously anticipate and align with others in real time. In contrast, existing works train humanoids as solitary agents, leaving them unable to perceive social cues or respond to the people around them. 
As humanoids move out of labs and into homes, workplaces, and shared public spaces, they will need the same capacity to operate alongside and with people.

To address this, we formulate the problem of human-humanoid interaction, in which a humanoid must continuously generate and execute coordinated, socially responsive whole-body motion given observations of a human partner.
We present \ourmethod{}, a hierarchical system with a high-level and a low-level policy. The high-level policy generates reactive humanoid motion from the observed partner's motion and anticipates the partner's motion, even when they are occluded or human input is missing, enabling continuous interaction. The low-level policy then executes the generated motion on a real Unitree G1 humanoid.

While recent advances in human motion generation~\cite{maluleke2024synergy,ghosh2024remos, cen2025ready, maluleke2025diffusionforcingmultiagentinteraction} can synthesize realistic dyadic interactions offline, deploying this capability on physical humanoids introduces three challenges:
First, a \emph{data gap}: human-human interaction is captured at scale, but no dataset captures natural human-humanoid interaction and naively retargeting each partner independently destroys the contact, reach, and spatial structure that defines the interaction. 
Second, a \emph{representation gap}: human and humanoid motion live in different spaces, which forces the generative model to learn the differences between the two embodiments instead of focusing on the interaction itself. This gap has been largely overlooked in prior work.
Third, a \emph{generation gap}: a natural approach is to generate the interaction with a human motion model and retarget to humanoid through optimization, but optimization-based retargeting is designed for clean, offline data and is brittle to noisy, discontinuous partner observations. A socially responsive humanoid needs a feed-forward model to generate humanoid motion directly.

\ourmethod{} addresses these challenges with three contributions that build on one another. (i) For the data gap, we introduce a two-phase Adaptive-HHI Retargeting procedure that retargets one partner to the 29-DoF Unitree G1 and morphs the other's body pose to preserve contact, reach, and relative scale. Applied to Chi3D~\cite{chi3dFieraru_2020_CVPR}, Hi4D~\cite{yin2023hi4d}, Harmony4D~\cite{khirodkar2024harmony4dvideodatasetinthewild}, Inter-X~\cite{xu2024inter}, and Embody3D~\cite{mclean2025embody3d}, it yields the HHI Benchmark, comprising 260 hours of dyadic and polyadic interactions paired with the G1. (ii) For the representation gap, our key insight is that a shared motion representation lets a single generative model reason jointly about both embodiments, which we instantiate with a shared-codebook VQ-VAE that tokenizes human and humanoid motion into a single discrete vocabulary. This vocabulary also supports real-time shadow teleoperation of the G1 without optimization-based retargeting (Sec.~\ref{sec:shadow_teleop}).. (iii) For the generation gap, our high-level policy is a Diffusion Forcing Transformer that autoregressively generates humanoid responses over this shared vocabulary; its diffusion-forcing objective is well-suited to imperfect, real-world input and supports fluid, long-horizon interaction. Paired with NVIDIA SONIC~\cite{luo2025sonic} as the low-level whole-body tracking controller, \ourmethod{} demonstrates coordinated, interactive, and human-aligned behaviors in both simulation and real-world hardware experiments. Generation and execution run asynchronously, with a median end-to-end latency of 308~ms from camera frame to humanoid command.

%% file: sec/02_related_work.tex
\begin{figure}[t]
    \centering
    \includegraphics[width=1\linewidth]{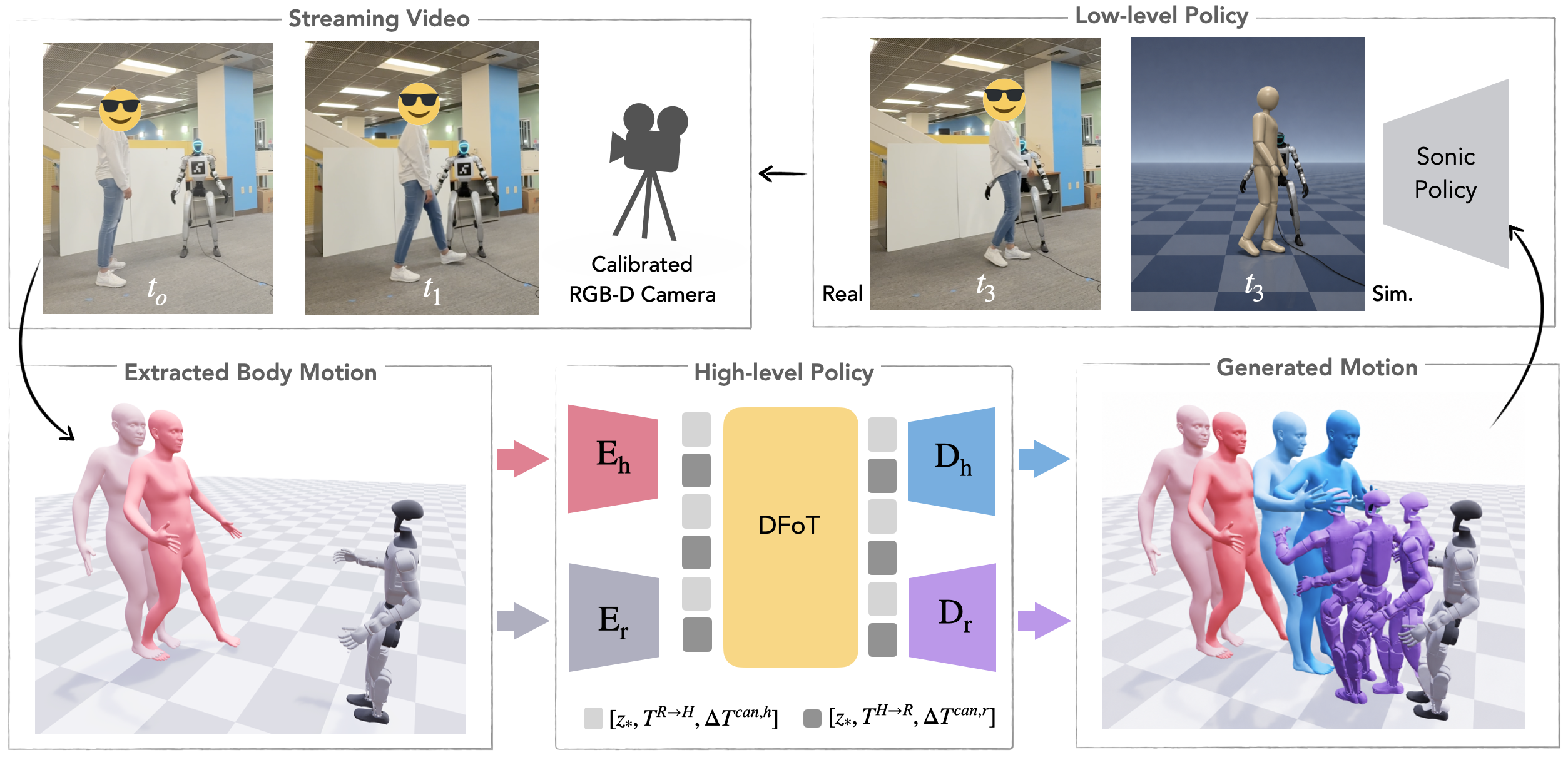}
    \caption{\textbf{\ourmethod{} Architecture.} From RGB-D observations of a human partner (red), we reconstruct the human pose and localize the body in 3D using depth and estimated gravity. Relative human and humanoid states are encoded through the VQ-VAE encoders $\mathrm{E}_{\mathrm{h}}$ and $\mathrm{E}_{\mathrm{r}}$. The resulting latent motion tokens $z_*$, relative pose transforms $T^*$, and canonical motion transforms $\Delta T^*$ are concatenated and processed by the DFoT policy, then decoded via $\mathrm{D}_h$ and $\mathrm{D}_r$ to future human behavior (blue) and socially coordinated humanoid motion (purple). SONIC executes the motion in sim or the real-world, and the robot state is updated for closed-loop interaction.}
    \label{fig:social_humanoid_architecture}
    \vspace{-1em}
\end{figure}

\section{Related Work}

\noindent\textbf{Human Motion Generation.}
Large-scale motion datasets such as AMASS~\cite{mahmood2019amass} have enabled major advances in generative human motion modeling. Early approaches relied on recurrent and convolutional architectures for motion prediction and synthesis~\cite{fragkiadaki2015recurrent,harvey2018recurrent,kaufmann2020convolutional,pavllo2018quaternetaq,habibie2017arv,barsoumcvprw2018}, followed by structured and probabilistic formulations for multimodal motion generation~\cite{aksan_2019_iccv,yuan2020dlow,zhang2020weam}. More recently, transformer-, VAE-, autoregressive-, and diffusion-based approaches have significantly improved long-horizon coherence, controllability, and diversity~\cite{duan2021single,guo2020action2motion,ahuja2019language2pose,jiang2023motiongpt,zhang2022motiondiffuse,zhang2023remodiffuse}.

\noindent\textbf{Humanoid Motion Generation.} Parallel progress in humanoid motion generation has combined reinforcement learning~\cite{yu2018learning,yu2019sim}, imitation learning and retargeting~\cite{choi2021self,Luo2023PerpetualHC,ze2025twistteleoperatedwholebodyimitation}, and generative motion priors~\cite{du2023avatars,zhang2024rohm,boige2023pasta,radosavovic2024humanoid,jiang2024harmon} to produce increasingly natural humanoid behaviors. Recent systems further leverage monocular video reconstruction and diffusion-based policies for contextual and goal-aware control~\cite{allshire2025visualimitationenablescontextual,liao2025beyondmimic}, while optimization-based methods such as OmniRetarget~\cite{omniretarget2025} improve geometric and kinematic consistency during motion transfer. However, these approaches primarily focus on single-agent motion generation or independently retargeted trajectories, often failing to preserve the contact, reachability, and relative geometry required for socially coordinated human-humanoid interaction. In contrast, our interaction-aware adaptive retargeting sequentially optimizes humanoid retargeting and partner co-morphing to preserve socially meaningful interaction structure.

\noindent\textbf{Dyadic Interaction and Human Motion Anticipation.}
Humans coordinate interactions by continuously anticipating others’ intentions through posture, timing, gaze, and motion cues~\cite{goleman2006social,freeman2016more,sebanz2006joint}. Motivated by this, recent work has explored dyadic human interaction~\cite{muller2024generative,fang2024capturing,wang2024intercontrol,ruiz2024in2in,ji2025towards,jiang2025arflow,cen2025ready,tanke2025dyadic,ghosh2025duetgen}, enabling realistic, coordinated behaviors between human partners. Complementary efforts study socially expressive and interaction-aware motion generation, including dance, social affordances, and language-conditioned interaction~\cite{liu2023interactive,siyao2024duolando,maluleke2024synergy,shan2024towards,fan2024freemotion}. While these methods produce compelling human-human interactions, they do not address the embodiment mismatch, streaming perception, and control constraints required for deployment on physical humanoids.

\noindent\textbf{Human-Humanoid Interaction and Embodied Control.}
Prior work on human-humanoid interaction has explored physical assistance and contact-rich behaviors such as standing support and hugging~\cite{lefevre2024humanoid,block2023arms}, as well as compliant and safe humanoid control for close-proximity interaction~\cite{margolis2025softmimic,lu2025gentlehumanoid}. Anticipating human intentions or trajectories has also been studied for cooperative robotics~\cite{huang2016anticipatory}, typically as auxiliary modules for timing or safety rather than as part of motion generation itself. Other recent approaches generate humanoid motion from high-level textual descriptions, including social interaction prompts~\cite{jiang2024harmon}. Hierarchical control frameworks further demonstrate the benefits of decomposing high-level motion planning and low-level motor execution for robust robot behavior~\cite{bajcsy2023learning}. In contrast to prior work, \ourmethod{} unifies predictive human motion understanding, cross-embodiment generative modeling, and embodied humanoid control into a single framework for socially coordinated human-humanoid interaction.

%% file: sec/03_method.tex
\section{Human-Humanoid Interaction Generative Modeling}

Enabling socially coordinated human-humanoid interaction requires not only realistic motion synthesis, but also the ability to react continuously to a human partner. We therefore train a generative model on human-humanoid interaction trajectories, enabling fast, reactive motion generation at inference time. The core of our \ourmethod{} system is a high-level policy that predicts humanoid kinematic motion from observed human motion, which are subsequently executed by a low-level tracking policy; see Fig.~\ref{fig:social_humanoid_architecture} for an overview of the system. The construction of the interaction dataset via interaction-aware retargeting is described in \Cref{sec:retarget}.

Formally, we adopt a canonical-frame motion representation rather than absolute world coordinates, inspired by MAGNet~\cite{maluleke2025diffusionforcingmultiagentinteraction}. Human articulated motion is given by SMPL joint rotations in 6D form, $\mathbf{\Theta}^h \in \mathbb{R}^{21 \times 6}$, and humanoid motion by the $29$ actuated joint angles $\mathbf{\Theta}^r \in \mathbb{R}^{29}$. Root motion is encoded as a relative canonical transform between consecutive frames, $\Delta \mathbf{T}^{\mathrm{can}} = [\Delta \mathbf{\Phi};\, \Delta \mathbf{\Gamma}]$, where $\Delta \mathbf{\Phi} \in \mathbb{R}^{6}$ is the relative root orientation (6D) and $\Delta \mathbf{\Gamma} \in \mathbb{R}^{3}$ the relative root translation in the canonical frame. We additionally retain the absolute canonical-to-root transform $\mathbf{T}^{\mathrm{can}\to\mathrm{root}}$ and the cross-agent transform $\mathbf{T}^{r\to h}$ ($\mathbf{T}^{h\to r}$) that places each agent in its partner's canonical frame, both in $\mathbb{R}^{9}$ (6D rotation, 3D translation). The complete per-agent motion representations are given by
\vspace{-2em}

\begin{equation}
\mathbf{m}^r =
[\,\Delta \mathbf{T}^{\mathrm{can},r};\,
\mathbf{T}^{\mathrm{can}\to\mathrm{root},r};\,
\mathbf{\Theta}^r;\,
\mathbf{T}^{r\to h}\,],
\hspace{8px}
\mathbf{m}^h =
[\,\Delta \mathbf{T}^{\mathrm{can},h};\,
\mathbf{T}^{\mathrm{can}\to\mathrm{root},h};\,
\mathbf{\Theta}^h;\,
\mathbf{T}^{h\to r}\,].
\label{eq:motion_rep}
\end{equation}

\subsection{Shared-Codebook VQ-VAE}

\label{sec:shared-vqvae}
\begin{figure}
    \centering
    \includegraphics[width=1\linewidth]{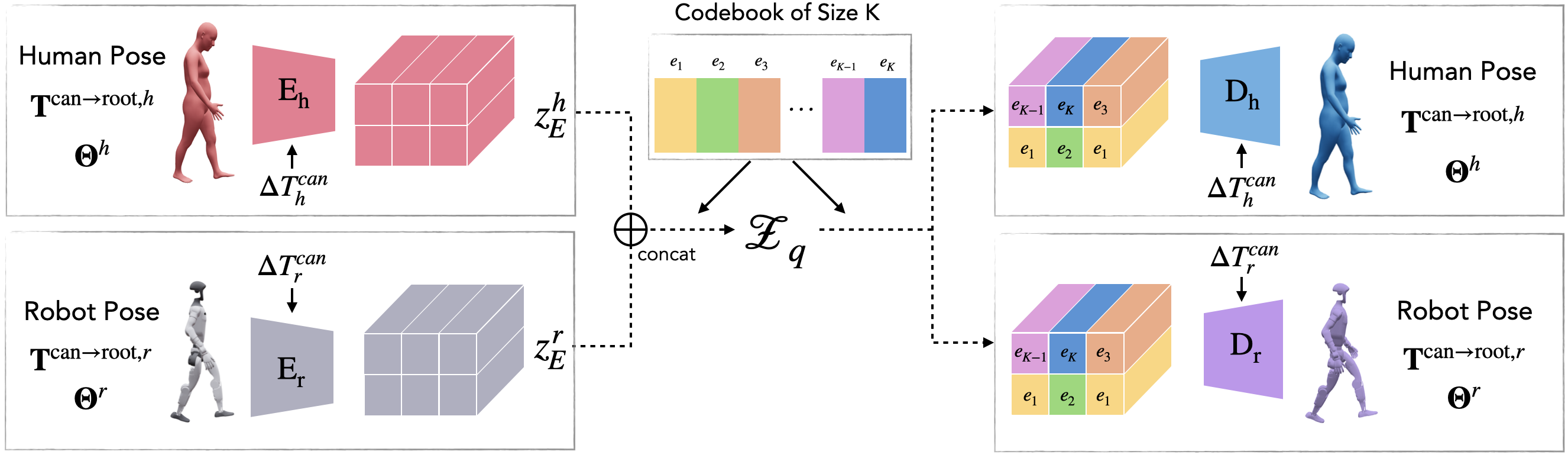}
    \caption{\textbf{Shared-Codebook VQ-VAE}. Human and humanoid motion sequences are independently encoded into latent features and mapped to a shared vector-quantized codebook that learns a common motion vocabulary across embodiments. Embodiment-specific decoders reconstruct the respective motions, enabling aligned, transferable motion representations between humans and humanoids.}
    \label{fig:shared_vqvae}
\end{figure}

Generating motion for Social Humanoids requires efficient, streaming-based generation, which motivates us to compactly tokenize high-dimensional human and humanoid motion trajectories. Recognizing that human and humanoid motions may share a similar latent structure, we encode both trajectories into a unified discrete vocabulary using a shared-codebook VQ-VAE (Fig.~\ref{fig:shared_vqvae}). Crucially, this shared representation simplifies the training of the subsequent diffusion model (see Sec.~\ref{sec:experiments}), as the primary motion spaces of the human partner and the humanoid are inherently heterogeneous.

\noindent\textbf{Embodiment-specific Encoders and Decoders.}
Human and robot motions lie in different feature spaces due to their differing dimensionalities and kinematic structures. We therefore use embodiment-specific encoders and decoders. Let $\mathbf{x}^{\mathrm{h}}_t =[\mathbf{T}_t^{\mathrm{can}\to\mathrm{root},h};\,
\mathbf{\Theta}_t^h] \in \mathbb{R}^{D_{\mathrm{h}}}$ and $\mathbf{x}^{\mathrm{r}}_t =[\mathbf{T}_t^{\mathrm{can}\to\mathrm{root},r};\,
\mathbf{\Theta}_t^r]  \in \mathbb{R}^{D_{\mathrm{r}}}$ denote the human and robot motion features at frame $t$. The human and robot encoders, $\mathrm{E}_{\mathrm{h}}$ and $\mathrm{E}_{\mathrm{r}}$, are 1D-convolutional networks with temporal downsampling by a factor of $4$ and dilated ResNet-1D blocks, producing latent sequences $\mathbf{v}^{\mathrm{h}}, \mathbf{v}^{\mathrm{r}} \in \mathbb{R}^{T_z \times d_z}$. 

These features are quantized into discrete sequences $\mathbf{z}^{\mathrm{h}}, \mathbf{z}^{\mathrm{r}}$ using a shared codebook. Finally, symmetric decoders, $\mathrm{D}_{\mathrm{h}}$ and $\mathrm{D}_{\mathrm{r}}$, upsample these quantized latents back to the original frame rate.

\noindent\textbf{Shared Codebook.} Both embodiments share a single EMA codebook, $\mathcal{C} = \{\mathbf{c}_k\}_{k=1}^{K} \subset \mathbb{R}^{d_z}$, defining a common discrete motion vocabulary. Given human and humanoid motions, the temporally downsampled continuous representations $\mathbf{v}^a$ produced by the encoders for agent $a \in \{\mathrm{h}, \mathrm{r}\}$ are quantized as:
\begin{equation}
\mathbf{z}^a = \mathrm{Quantize}(\mathbf{v}^a; \mathcal{C}), \qquad \text{where} \quad \mathbf{z}^a_t = \arg\min_{\mathbf{c}_k \in \mathcal{C}} \|\mathbf{v}^a_t - \mathbf{c}_k\|_2.
\end{equation}
The resulting discrete latents $\mathbf{z}^a$ are subsequently passed to their respective decoders.

However, a shared codebook alone cannot guarantee identical code assignments for paired frames. We thus introduce two alignment terms:
\textbf{(i) Continuous alignment} directly minimizes the distance between encoder outputs, $\mathcal{L}_{\mathrm{align}} = \|\mathbf{v}^{\mathrm{h}} - \mathbf{v}^{\mathrm{r}}\|_2^2$, shaping a shared latent geometry. 
\textbf{(ii) Discrete alignment} applies a symmetric cross-entropy loss over the distance logits $\ell^{\mathrm{h}}, \ell^{\mathrm{r}} \in \mathbb{R}^{K}$ with stop-gradient targets $c^{\mathrm{h}}, c^{\mathrm{r}}$, computed as $\mathcal{L}_{\mathrm{agree}} = \tfrac{1}{2} [ \mathrm{CE}(\ell^{\mathrm{h}}, c^{\mathrm{r}}) + \mathrm{CE}(\ell^{\mathrm{r}}, c^{\mathrm{h}}) ]$. 
Together, these terms encourage cross-embodiment code prediction and tie the codebook to corresponding human--robot poses rather than letting each encoder claim separate regions of $\mathcal{C}$.

\noindent\textbf{Training Objective.} The complete objective combines per-embodiment reconstruction, the VQ commitment loss $\mathcal{L}_{\mathrm{commit}} = \|\mathbf{v} - \mathrm{sg}(\mathbf{z})\|_2^2$, and the two proposed alignment terms:
\begin{equation}
\mathcal{L} = \mathcal{L}_{\mathrm{rec}} + \beta\,\mathcal{L}_{\mathrm{commit}} + \lambda_{align}\,\mathcal{L}_{\mathrm{align}} + \lambda_{agree}\,\mathcal{L}_{\mathrm{agree}}.
\label{eq:vqvae_training_obj}
\end{equation}
The codebook entries are the only shared parameters across the architecture; all other model weights remain strictly embodiment-specific. The shared codebook also enables human-to-humanoid motion translation in a single feed-forward pass (\(E_h \rightarrow \mathcal{C} \rightarrow D_r\)), supporting real-time shadow teleoperation of the G1 without optimization-based retargeting (Appendix~\ref{sec:shadow_teleop}).

\subsection{Sequence Modeling of Human--Humanoid Interaction} 

To model human–humanoid interaction in the shared token space introduced above, we use a flexible sequence model based on Diffusion Forcing Transformer (DFoT)~\cite{chen2025diffusion, song2025historyguidedvideodiffusion}. DFoT is a conditional generative model that integrates the denoising process of DDPMs \cite{ho2020denoising} with the temporal strengths of transformers \cite{vaswani2017attention}. A core advantage of this design is its ability to learn non-Markovian temporal dependencies across entire sequences using adaptive noise levels.

Given the observed latent history $\{\mathbf{z}^{\mathrm{h}}, \mathbf{z}^{\mathrm{r}}\}_{1:T_c}$, DFoT predicts the future latents $\{\hat{\mathbf{z}}^{\mathrm{h}}, \hat{\mathbf{z}}^{\mathrm{r}}\}_{T_c+1:T_c+T_f}$. During training, each future latent frame $\mathbf{z}^a_i$ for agent $a \in \{\mathrm{h}, \mathrm{r}\}$ at timestep $i$ is noised independently according to its own frame- and agent-specific diffusion timestep $t^a_i$, yielding 
$
\mathbf{z}^a_i(t^a_i) = \sqrt{\bar{\alpha}_{t^a_i}}\, \mathbf{z}^a_i + \sqrt{1 - \bar{\alpha}_{t^a_i}}\, \boldsymbol{\epsilon}^a_i,
$
where $\mathbf{z}^a_i$ is the clean latent, $\bar{\alpha}$ is the noise schedule, and $\boldsymbol{\epsilon}^a_i \sim \mathcal{N}(\mathbf{0}, \mathbf{I})$. Crucially, human and robot frames are noised independently ($t^{\mathrm{h}}_i \neq t^{\mathrm{r}}_i$) to prevent shortcutting. Context frames $1{:}T_c$ remain clean ($t^a_i = 0$) to preserve the history conditioning signal.

The network consists of transformer layers processing the noisy latents $\mathbf{z}^a_i(t^a_i)$ combined with a sinusoidal embedding for noise-level, RoPE for temporal order, and a learned agent embedding. We train with a Smooth-$L_1$ objective to recover clean latents, and during inference, the predicted future latents are passed to embodiment-specific decoders for motion reconstruction.

\noindent\textbf{Sampling and Generation.}
Inference with DFoT uses autoregressive DDIM sampling~\cite{song2020denoising} with $\eta=0$ to generate coordinated human and humanoid motion. The denoising process is controlled by a sampling strategy matrix $\mathbf{K} \in \mathbb{Z}^{S \times (N \times P)}$, which specifies per-token noise levels across denoising steps. Independent token-wise noise scheduling enables flexible generation over the agent--time grid, allowing the same model to perform tasks such as joint future generation, partner prediction, and partner inpainting through different sampling schedules without retraining.

\noindent\textbf{Training Losses.} We optimize \ourmethod{} across three complementary spaces---latent, motion, and reconstruction---to balance diffusion accuracy with kinematic fidelity:
\begin{equation}
\mathcal{L}_{\text{total}} = w_{\text{latent}} \mathcal{L}_{\text{latent}} + w_{\text{motion}} \mathcal{L}_{\text{motion}} + w_{\text{recon}} \mathcal{L}_{\text{recon}},
\end{equation}
where $w_{\text{latent}}$, $w_{\text{motion}}$, and $w_{\text{recon}}$ are weighting coefficients.
The \emph{latent} denoising loss recovers clean latents via Smooth-$L_1$ for human ($h$) and robot ($r$); the \emph{motion} loss decodes predicted latents and applies Smooth-$L_1$ in raw motion space against $\mathbf{x}^a$ from Eq.~1; the \emph{reconstruction} loss enforces a robust bottleneck by passing
ground-truth motion through a non-diffused autoencoding loop.

\subsection{End-to-End Deployment}
\label{sec:e2e}

We deploy \ourmethod{} on a Unitree G1 using the perception, generation, and control pipeline in Fig.~\ref{fig:social_humanoid_architecture}, with most computation offboard. A tripod-mounted ZED 2i stereo camera observes the human partner. We estimate the partner's SMPL pose using Fast SAM 3D Body~\cite{yang2026fastsam3dbody} and recover metric root position from stereo depth. AprilTags and onboard odometry localize the humanoid in the same gravity-aligned frame. We encode both agents' states into the shared interaction representation for DFoT, which autoregressively generates humanoid motion conditioned on observed partner motion and recent humanoid history using 30-step DDIM sampling. NVIDIA SONIC~\cite{luo2025sonic} tracks the decoded motion at 50~Hz. Generation and execution run asynchronously, with 308~ms median latency from camera frame to executed humanoid command. See Appendix~\ref{app:system} for details.

\section{Humanoid-Human Interaction Data Generation}
\label{sec:retarget}

Social humanoid development is bottlenecked by a lack of human-humanoid interaction datasets. While human-human data is abundant, direct kinematic retargeting fails due to morphology gaps. To this end, we propose an interaction-preserving retargeting method maintaining contact locations, relative poses, and inter-person geometry. 

\begin{figure}
    \centering
    \includegraphics[width=1\linewidth]{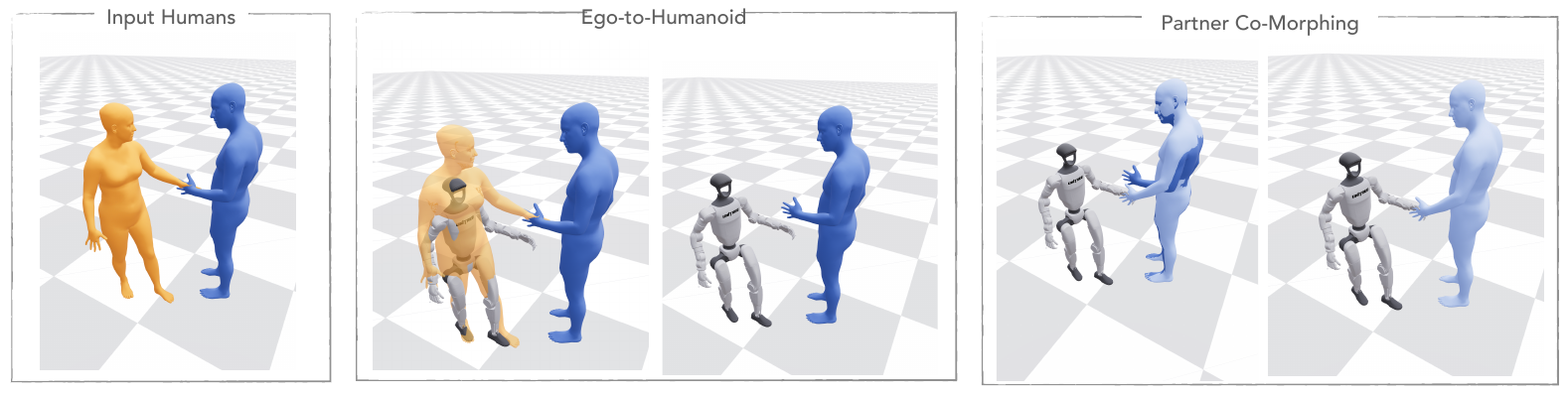}
    \caption{\textbf{Adaptive-HHI Retargeting.} Our retargeting pipeline starts from two ground truth humans. In the first pass, we retarget the ego human (orange) to the humanoid using a contact-adaptive weight matrix to encourage body pose in regions close to the physical interaction to be preserved. In a second pass, we co-morph the partner's pose towards more plausible interaction via a contact-gated morphing strategy driven by the amount of physical contact present in the interaction. }  
    \label{fig:retarget}
    \vspace{-1.3em}
\end{figure}

Building on \textsc{OmniRetarget}~\cite{omniretarget2025}'s Sequential Quadratic Programming (SQP) and interaction-mesh formulation~\cite{ho2010spatial}, we address a central challenge of dyadic retargeting: bridging the morphology gap without distorting the interaction. Retargeting the humanoid independently can place interaction points out of reach, while uniformly rescaling the pair would make the human partner unrealistically small, creating a mismatch between the people the humanoid interacts with at deployment. We use a two-pass pipeline that retargets the ego, then adapts the partner's pose—not its scale—to the retargeted humanoid (Fig.~\ref{fig:retarget}).

\noindent\textbf{Pass~I: Ego-to-humanoid Retargeting.}
Given an interaction between a source human (the ego) and a partner, this pass retargets the ego to the humanoid, leaving the partner unchanged. Following \textsc{OmniRetarget}~\cite{ho2010spatial,omniretarget2025}, we solve a per-frame SQP over the humanoid configuration that preserves the local shape of the ego's interaction mesh, with Laplacian coordinates scaled to humanoid height, and keeps the humanoid's wrist and elbow distances to the partner equal to the ego's. Temporal smoothness and posture regularization complete the objective, subject to joint limits and foot-sticking constraints. Both the shape and distance terms emphasize the wrists, where most contact occurs. The full objective is given in Appendix~\ref{supp:passI} and our contact-aware grounding in Appendix~\ref{supp:contact_aware_grounding}.

\noindent\textbf{Pass~II: Partner Co-Morphing.} 
Pass~I leaves the human partner unchanged, causing contact misalignment because the G1 is substantially shorter than the source human. For example, the partner's hands may reach above the humanoid's shoulders during hugs. Pass~II adapts the partner's pose to the retargeted humanoid. 
For all contacts, we apply arm-chain IK over the partner's shoulder, elbow, and wrist during frames with humanoid arm contact, preserving the source ego--partner wrist and elbow offsets relative to the retargeted humanoid while respecting anatomical joint limits. For high-contact segments, we additionally solve a body-pose QP over key vertex clusters (wrists, chest, and neck), which additionally adjusts the partner's hips, spine, and knees. This restores contact fidelity across large morphology gaps while preserving realistic human proportions. 

We retarget each clip twice, swapping the ego and partner, so both interaction roles are represented in the dataset.

%% file: sec/04_experiments.tex
\section{Experiments}
\label{sec:experiments}

\minisection{Metrics} We evaluate three tasks. For generative modeling, given the past and current human motion and the past humanoid motion, the high-level policy predicts the humanoid's reactive motion, which we compare against the retargeted HHI data in terms of positional accuracy, motion dynamics, physical realism, and interaction coordination. For retargeting, we evaluate how faithfully a human--human interaction is transferred to the humanoid. Finally, we evaluate whether the low-level policy can execute the generated motions in simulation and on the real humanoid.

\textit{W-MPJPE} is the mean per-joint 3D position error in the world frame. \textit{MPJVE} is the mean per-joint velocity error, which captures temporal smoothness. \textit{Foot Skating (FS)} is the skating contact ratio, measuring unrealistic foot sliding. \textit{Interpenetration (IP)} is the penetration depth between human and humanoid body parts. \textit{Motion Interaction (MI)} is the difference between the ground-truth and predicted correlations of human and humanoid joint positions. \textit{Fréchet Distance (FD)} compares the distributions of real and generated motion to measure overall quality with respect to the training data. For retargeting, \textit{Pose distortion} is the root-relative MPJPE (cm) between the humanoid links and the corresponding joints of the human, scaled to humanoid height. $E_{\text{wrist}}$ (cm) is the mean absolute error between the humanoid--partner wrist distance and the demonstrated ego--partner wrist distance, where partner joints are taken at the homologous SMPL-X joint and the demonstration is scaled to humanoid height. Lower is better for all metrics except contact preservation and success rate.

\minisection{Baselines} 
\label{sec:baseline}
Evaluating socially responsive humanoid interaction is challenging because existing methods either synthesize human-human interactions offline or generate isolated humanoid behaviors without addressing executable cross-embodiment interaction generation. We therefore compare against retrieval-based Nearest Neighbor (NN) and Random Sample (RS), which measure dataset memorization and distributional coverage, and against Ready-to-React (R2R)~\cite{cen2025ready}, a reaction policy for two-character interaction generation. NN retrieves the training window with the lowest L2 distance between normalized partner joint positions in the canonical frame, and RS samples a window uniformly at random.

\newlength{\tablegap}
\setlength{\tablegap}{1em} 
\begin{table}[t]
\centering
\scriptsize
\setlength{\tabcolsep}{2pt}
\renewcommand{\arraystretch}{1.1}
\begin{minipage}[t]{\dimexpr(\linewidth-\tablegap)/2\relax}
\centering
\captionof{table}{\textbf{Baseline comparison on HHI data.}}
\label{tab:baseline_comparison}
\begin{tabular*}{\linewidth}{@{}l@{\extracolsep{\fill}}cccccc@{}}
\toprule
 & Qual. & \multicolumn{3}{c}{Phys. Realism} & Dyn. & Pos. \\
\cmidrule(lr){2-2} \cmidrule(lr){3-5} \cmidrule(lr){6-6} \cmidrule(l){7-7}
\textbf{Method} & FD$\downarrow$ & FS$\downarrow$ & IP$\downarrow$ & MI$\downarrow$ & MPJVE$\downarrow$ & W-MPJPE$\downarrow$ \\
\midrule
NN    & \textbf{0.053} & 0.138 & 0.057 & 0.056 & 0.014 & 1.099 \\
RS    & 0.180 & \textbf{0.134} & 0.058 & 0.056 & 0.015 & 1.100 \\
R2R \cite{cen2025ready}   & 0.504 & 0.921 & 0.088 & 0.088 & 0.019 & 0.545 \\
\midrule
\fullrowcolor{gray!12}
\textbf{Ours} & 0.118 & 0.876 & \textbf{0.051} & \textbf{0.002} & \textbf{0.007} & \textbf{0.128} \\
\bottomrule
\end{tabular*}
\end{minipage}%
\hspace{\tablegap}%
\begin{minipage}[t]{\dimexpr(\linewidth-\tablegap)/2\relax}
\centering
\captionof{table}{\textbf{Ablations on HHI-Chi3D.}}
\label{tab:ablations_combined}
\begin{tabular*}{\linewidth}{@{}l@{\extracolsep{\fill}}cccccc@{}}
\toprule
 & Qual. & \multicolumn{3}{c}{Phys. Realism} & Dyn. & Pos. \\
\cmidrule(lr){2-2} \cmidrule(lr){3-5} \cmidrule(lr){6-6} \cmidrule(l){7-7}
\textbf{Variant} & FD$\downarrow$ & FS$\downarrow$ & IP$\downarrow$ & MI$\downarrow$ & MPJVE$\downarrow$ & W-MPJPE$\downarrow$ \\
\midrule
No VQ-VAE   & 0.253 & 0.931 & 0.027 & \textbf{0.018} & 0.007 & 0.075 \\
Sep. VQ-VAE & 0.181 & \textbf{0.833} & 0.088 & 0.212 & 0.007 & 0.085 \\
\midrule
\fullrowcolor{gray!12}
\textbf{Ours} & \textbf{0.072} & 0.845 & \textbf{0.026} & 0.019 & \textbf{0.005} & \textbf{0.057} \\
\bottomrule
\end{tabular*}
\end{minipage}
\end{table}

\subsection{Results} 
\label{sec:exp:motion}

In Table~\ref{tab:baseline_comparison}, we present a comprehensive comparison across various evaluation metrics assessing motion and interaction quality, plausibility, and fidelity of the high-level policy on our HHI dataset.

\noindent \textbf{Retargeting Motion Quality.} Table~\ref{tab:multidataset-retarget-comparison} compares our retargeting method, Adaptive-HHI Retargeting, with PHC and OmniRetarget on CHI3D, Hi4D, and Harmony4D. As no ground-truth humanoid motion exists, we evaluate pose distortion of each method against the source human--human demonstration, scaled to the humanoid's height. Ours achieves the lowest pose distortion, highest contact preservation, and lowest foot skating on all three datasets. The largest gain is in interaction fidelity: wrist error $E_{\text{wrist}}$ falls from 1.38 to 0.50\,cm on CHI3D and from 2.98 to 1.91\,cm on Harmony4D relative to OmniRetarget, while PHC exceeds 20\,cm. For the downstream RL policy, OmniRetarget and ours both reach 100\% success on CHI3D and Hi4D, but on Harmony4D ours reaches 97\% versus 81\% for OmniRetarget and 62\% for PHC.

\begin{table}[t]
\centering
\caption{
\textbf{Retargeting comparison across three interaction datasets.} Best per column in bold; $\uparrow$/$\downarrow$ indicate higher/lower is better.
}
\label{tab:multidataset-retarget-comparison}
\small
\setlength{\tabcolsep}{4pt}

\resizebox{\linewidth}{!}{%
\begin{tabular}{llccccc}
\toprule
& & \textbf{Pose Distortion}
& \textbf{Contact Preservation}
& \textbf{Foot Skating}
& \textbf{Interaction}
& \textbf{Downstream RL Policy} \\
\cmidrule(lr){3-3}
\cmidrule(lr){4-4}
\cmidrule(lr){5-5}
\cmidrule(lr){6-6}
\cmidrule(lr){7-7}
\textbf{Dataset}
& \textbf{Method}
& MPJPE [cm]~$\downarrow$
& Dur.~$\uparrow$
& Dur.~$\downarrow$
& $E_{\text{wrist}}$ [cm]~$\downarrow$
& Success Rate(\%)~$\uparrow$ \\
\midrule

\multirow{3}{*}{CHI3D}
& PHC      & 6.179 & 0.478 & 0.107 & 22.382 & 68  \\
& OmniRetarget & 4.184 & 0.915 & 0.009 & 1.379  & \best{100} \\
\rowcolor{lightgray}
& Ours         & \best{3.893} & \best{0.936} & \best{0.003} & \best{0.501} & \best{100} \\
\midrule

\multirow{3}{*}{Hi4D}
& PHC          & 6.892 & 0.089 & 0.087 & 21.088 & 79  \\
& OmniRetarget & 4.704 & 0.827 & 0.006 & 1.729  & \best{100} \\
\rowcolor{lightgray}
& Ours         & \best{4.371} & \best{0.881} & \best{0.005} & \best{0.804} & \best{100} \\
\midrule

\multirow{3}{*}{Harmony4D}
& PHC          & 7.656 & 0.215 & 0.291 & 23.365 & 62 \\
& OmniRetarget & 5.220 & 0.676 & 0.033 & 2.981  & 81 \\
\rowcolor{lightgray}
& Ours         & \best{4.638} & \best{0.775} & \best{0.020} & \best{1.909} & \best{97} \\
\bottomrule
\end{tabular}%
}
\vspace{-8px}
\end{table}

\begin{figure}[th]
    \centering
    \includegraphics[width=0.9\linewidth]{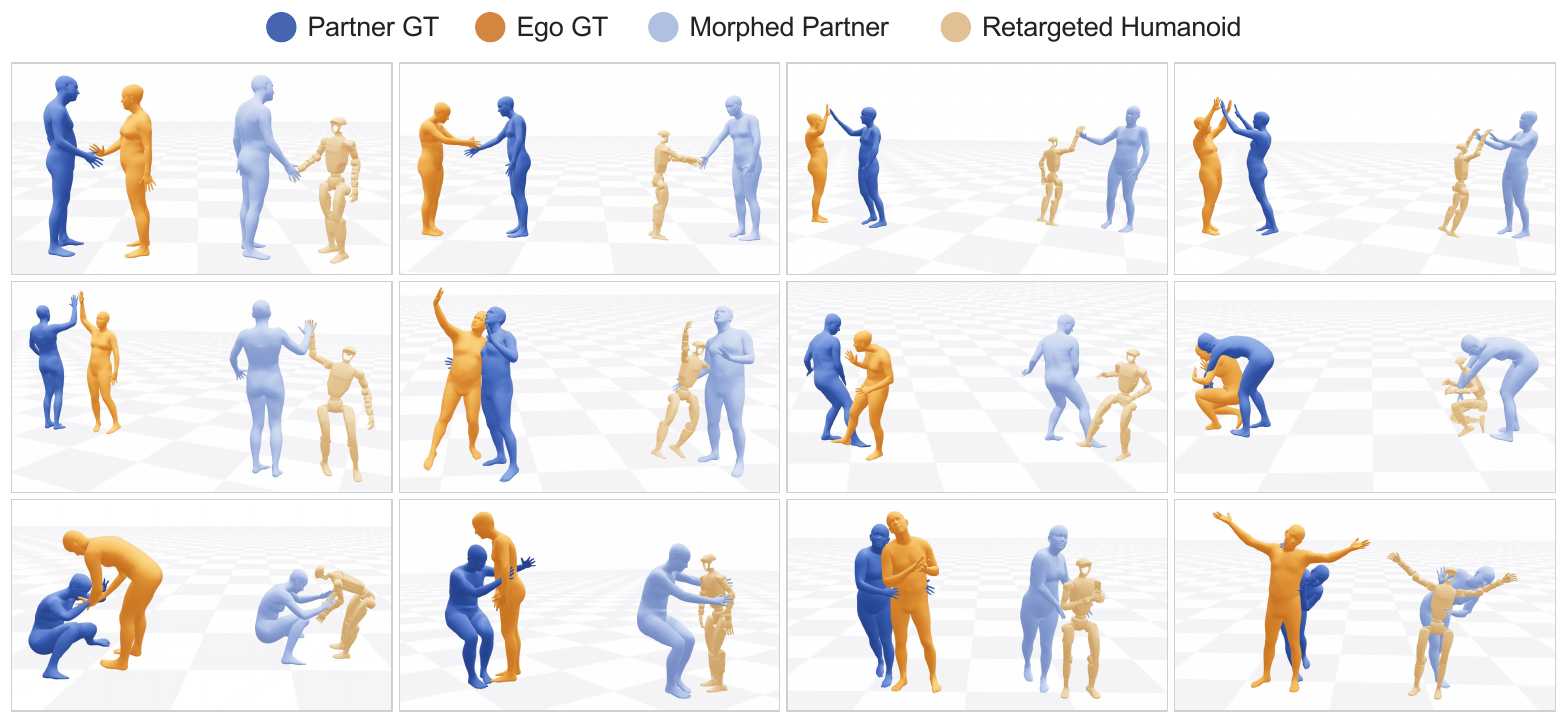}
    \caption{\textbf{Retargeting Qualitative Results.} Each panel shows a ground-truth human--human interaction (ego in orange, partner in dark blue) next to its retargeted counterpart. The ego is retargeted to the humanoid (tan), and the partner's pose is morphed to preserve interaction (light blue). Across diverse interactions our Adaptive-HHI Retargeting method preserves the contacts and relative positioning between the two agents.}
    \label{fig:retargeting_results}
\end{figure}

\begin{figure}
    \centering
    \includegraphics[width=1\linewidth]{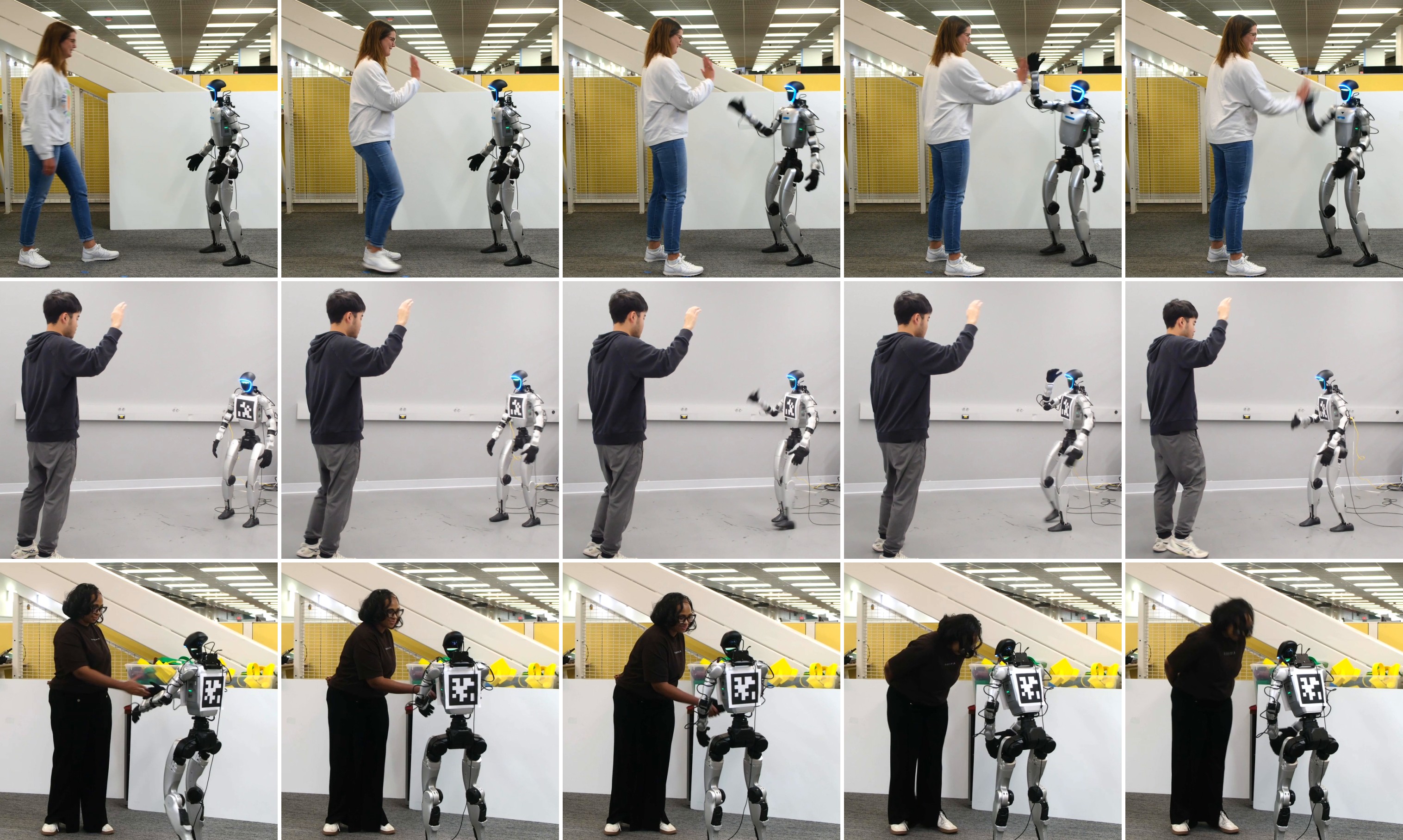}
    \caption{\textbf{Real-world deployment}. Frames run left to right in time. Top two rows: separate high-five and hand-wave trials. Bottom: a single continuous take in which the humanoid transitions directly from a handshake into a bend}
    \label{fig:RealDemo}
\end{figure}

\noindent\textbf{High-level Policy Motion Quality.} Table~\ref{tab:baseline_comparison} evaluates the generative modeling capability of the high-level policy. Our method achieves the lowest positional error with a W-MPJPE of 0.128, substantially outperforming R2R (0.545), NN (1.099), and RS (1.100), and the best motion dynamics with an MPJVE of 0.007, indicating smoother, more temporally coherent motion. It also achieves the best interpenetration (0.051) and motion interaction (0.002), indicating physically consistent interactions with the human partner. NN achieves the best FD (0.053 vs.\ our 0.118), and NN and RS have lower foot skating (FS), since retrieval replays recorded training motions with physically valid foot contacts. Despite this limitation, the proposed approach still produces coherent full-body motions with substantially improved interaction.

\noindent\textbf{Ablations.} In~\Cref{tab:ablations_combined}, we ablate two design choices of the high-level generative model on HHI-Chi3D to assess the importance of discrete motion tokenization and codebook sharing: removing VQ-VAE tokenization (No VQ-VAE) and using per-agent rather than shared codebooks with no alignment (Sep.\ VQ-VAE). Our shared-codebook model achieves the best FD (\textbf{0.072}), IP (\textbf{0.026}), MPJVE (\textbf{0.005}), and W-MPJPE (\textbf{0.057}), confirming that discrete tokenization regularizes the high-level policy and a shared codebook is key to coherent interaction.

\begin{wraptable}[11]{R}{0.3\columnwidth}
\vspace{-1.2em}
\caption{\textbf{Success rate (\%) in sim and on the real G1.}}
\label{tab:sim_robot_results}
\centering
\footnotesize
\setlength{\tabcolsep}{3pt}
\renewcommand{\arraystretch}{1.05}
\begin{tabular}{@{}lcc@{}}
\toprule
\textbf{Scenario} & \textbf{Sim} & \textbf{Real} \\
\midrule
Handshake     & 100 & 100 \\
Hand wave     & 100 & 100 \\
High-five     & 90  & 90  \\
Hug           & 100 & 100 \\
Bend          & 100 & 100 \\
Linked arms   & 100 & 100 \\
\bottomrule
\end{tabular}
\end{wraptable}

\noindent\textbf{Sim and Real Humanoid Evaluation.} Our retargeting pipeline and high-level policy do not model physics explicitly, so we test whether the generated humanoid motions are physically feasible by executing them with the low-level policy, both in simulation and on the real G1. In simulation, a rollout succeeds if the low-level policy tracks the generated motion without falling or other unsafe behavior; we evaluate 30 generated motions per scenario. On the real humanoid, a trial succeeds if the G1 executes the appropriate reaction and completes the interaction without falling or operator intervention; we run 30 trials per scenario across 4 participants. As Table~\ref{tab:sim_robot_results} shows, the low-level policy reproduces nearly all generated motions in simulation (90--100\% per scenario), and on the real G1 (Fig.~\ref{fig:RealDemo}) success matches simulation in every scenario.

\subsection{Shadow Teleoperation}
\label{sec:shadow_teleop}

Beyond interaction generation, the shared-codebook VQ-VAE supports shadow teleoperation, in which the humanoid reproduces the whole-body motion of a human operator. As in cross-embodiment translation (Appendix~\ref{app:Cross-Embodiment Translation}), we encode the operator's motion with the human encoder $E_h$, quantize it against the shared codebook $\mathcal{C}$, and decode the resulting tokens with the humanoid decoder $D_r$. No humanoid motion is used as input, and no optimization-based retargeting is run.

\textbf{Role of the shared codebook.} Shadow teleoperation relies on human and humanoid poses mapping to the same codes, which is enforced by the alignment terms $\mathcal{L}_{\text{align}}$ and $\mathcal{L}_{\text{agree}}$ in Eq.~\ref{eq:vqvae_training_obj}. Without them, the shared codebook reconstructs each embodiment well but translates poorly, with a human-to-humanoid MPJPE of 12.52\,cm; with both terms, this error falls to 2.79\,cm and code agreement between paired frames rises from 0.00 to 0.78 (Table~\ref{tab:ablation-vqvae}).

\begin{figure}
    \centering
    \includegraphics[width=1\linewidth]{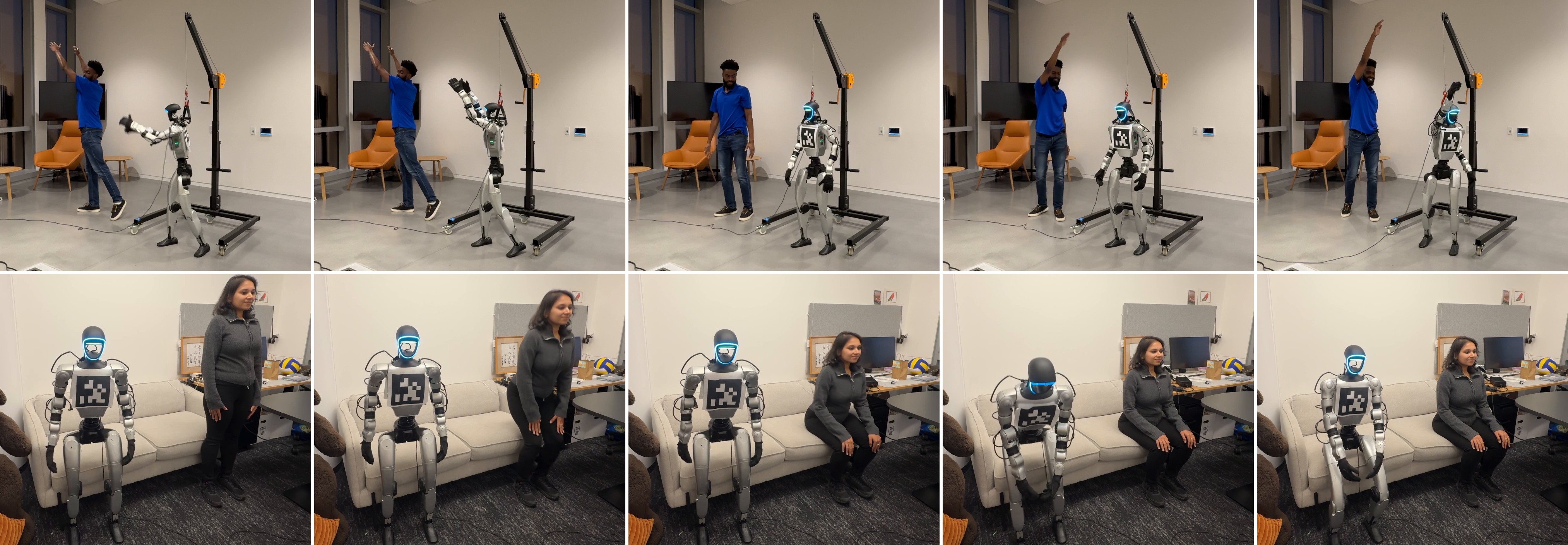}
    \caption{ \textbf{Shadow teleoperation}. The G1 shadows a human operator performing arm movements (top) and sitting down on a couch (bottom). The operator's motion is encoded with $E_h$, quantized against the shared codebook $\mathcal{C}$, and decoded with $D_r$ into humanoid motion. Frames run left to right in time.}
    \label{fig:shadow_teleop}
    \vspace{-1.5em}
\end{figure}

\textbf{Qualitative results.} Fig~\ref{fig:shadow_teleop} shows the G1 shadowing an operator who sits down on a couch. As the operator moves from standing to sitting, the humanoid follows into a seated posture through its own kinematics rather than copying joint angles one-to-one.

%% file: sec/05_conclusion.tex
\section{Conclusion}

We present \ourmethod{}, a system for generating and executing socially responsive whole-body motion on a Unitree G1 conditioned on a human partner. It combines a two-phase Adaptive-HHI Retargeting procedure that produces paired human--humanoid data while preserving contact and scale, a shared-codebook VQ-VAE that maps both embodiments into a single discrete latent space, which also enables real-time shadow teleoperation, and a Diffusion Forcing Transformer that generates humanoid responses in this space, executed on the robot via SONIC. Together with the 260-hour HHI Benchmark, we show in simulation and on the real G1 that the system produces coordinated, human-aligned responses across a range of interaction types. We will release our data and code to support further work toward humanoids that coordinate physically with people.

\minisection{Limitations and Future Work} First, we do not model hand articulation, facial expressions, or object pose, limiting dexterous, expressive, and object-centric interactions. Extending the shared codebook to fine-grained hand motion is a natural next step, while object interaction will likely require additional representation design. Second, we rely on SONIC and off-the-shelf human pose estimation, and inherit their failure modes in real-world deployment. Our system also tracks kinematic trajectories without modeling interaction forces, human comfort, or safety constraints during close contact, so safe deployment in unconstrained environments will require additional safety policies. Third, trackability and trial success measure execution, not social interaction quality; an HRI study is needed to assess the timeliness, naturalness, safety, and appropriateness of the humanoid's responses. Fourth, we focus on dyadic human–humanoid interaction; extending to multiple humans, robots, or heterogeneous agents is an important direction for future work. Finally, when interacting with the robot, the lack of facial expressions or eye contact makes its behavior difficult to interpret, leading to awkward social interactions. An interesting direction for future work would be to enable more legible social interactions through facial expressions and eye contact.

%% file: sec/supp.tex
\clearpage
\renewcommand{\thepart}{}
\renewcommand{\partname}{}
\doparttoc
\faketableofcontents

\addcontentsline{toc}{section}{Appendix}
\part{Supplementary Material}
\parttoc

\renewcommand{\thetable}{A.\arabic{table}}
\setcounter{table}{0} 
\renewcommand{\thesection}{A.\arabic{section}}
\setcounter{section}{0}  
\renewcommand{\thefigure}{A.\arabic{figure}}
\setcounter{figure}{0}

\section{Introduction}
We provide additional details on the dataset (Section~\ref{app:dataset}), retargeting (Section~\ref{app:retargeting}), high-level policy (Section~\ref{app:high-level}), and end-to-end system (Section~\ref{app:system}) for \textit{Social Humanoid}.

\section{Human-Humanoid Interaction (HHI) Dataset}
\label{app:dataset}

\begin{table}[h]
\centering
\small
\caption{\textbf{Summary statistics of our retargeted benchmark.} Clip counts, frames, and durations are reported after retargeting each source dataset to the Unitree G1 with Adaptive-HHI.}
\label{tab:datasets_summary}
\begin{tabular}{lrrr}
\toprule
\textbf{Dataset} & \textbf{Clips} & \textbf{Total Frames} & \textbf{Total Time (hrs)} \\
\midrule
CHI3D \cite{chi3dFieraru_2020_CVPR}                            & 245    & 25,588     & 0.24   \\
Hi4D \cite{yin2023hi4d}                                        & 200    & 23,890     & 0.22   \\
Harmony4D \cite{khirodkar2024harmony4dvideodatasetinthewild}  & 416    & 153,160    & 1.42   \\
Inter-X \cite{xu2024inter}                                     & 22,776 & 4,049,990  & 37.50  \\
Embody 3D \cite{mclean2025embody3d}                            & 14,692 & 23,864,668 & 220.97 \\
\bottomrule
\end{tabular}
\end{table}

Table~\ref{tab:datasets_summary} summarizes our unified human-humanoid Interaction (HHI) benchmark, constructed by retargeting five popularly used human-human interaction mocap datasets to the Unitree G1 humanoid with our Adaptive-HHI pipeline (Sec.~\ref{app:retargeting}). The source datasets span a diverse spectrum of close-range interactive behaviors --- from contact-rich dyadic interactions (Chi3D~\cite{chi3dFieraru_2020_CVPR}, Hi4D~\cite{yin2023hi4d}) and in-the-wild multi-person activities (Harmony4D~\cite{khirodkar2024harmony4dvideodatasetinthewild}) to large-scale captures of multi-person interactions (Inter-X~\cite{xu2024inter}, Embody~3D~\cite{mclean2025embody3d}). Each clip is processed according to its number of participants, and all sequences are resampled to 30\,FPS. We adopt an 80/10/10\% train/validation/test split.

\section{Retargeting Details}
\label{app:retargeting}

In this section, we compare Adaptive-HHI to existing retargeting methods and provide additional implementation details and hyperparameter configurations for our retargeting pipeline.

Table~\ref{tab:retarget-comparison} compares Adaptive-HHI to prior retargeting methods along four axes: hard kinematic feasibility, preservation of inter-agent interaction, partner co-morphing, and adaptive contact handling. Interaction meshes were introduced for close-interaction character animation~\cite{ho2010spatial} and later adapted to biped humanoids~\cite{imma2012}, where Laplacian deformation preserves spatial relationships under hard joint constraints, but only for the robot's own body parts; the demonstrated interactions are self-contacts, and no human partner enters the optimization. PHC~\cite{Luo2023PerpetualHC} imitates single-agent motion through a simulated physics policy, so feasibility is implicit and retargeted references can still exhibit motion artifacts. GMR~\cite{ze2025twistteleoperatedwholebodyimitation} solves per-frame IK with joint and velocity limits, but its soft keypoint-matching objectives admit foot skating and penetration, and it has no notion of a partner. VideoMimic~\cite{allshire2025visualimitationenablescontextual} adds scene context but handles collision and contact through soft penalties that compete with its matching costs. OmniRetarget~\cite{omniretarget2025} formulates retargeting as an interaction-mesh-regularized SOCP with hard constraints and preserves robot--object--terrain relationships, yet the surrounding scene is a fixed input to the solve, not an optimization variable. This is the core limitation for human--humanoid interaction: when the partner cannot adapt, the robot must absorb the entire embodiment gap, and the interaction breaks. Adaptive-HHI keeps OmniRetarget's SOCP base and adds (i) partner co-morphing, adapting the partner's arm configuration to the retargeted humanoid on contact frames so the dyad deforms together, and (ii) wrist- and elbow-distance preservation with wrist-boosted Laplacian weights that concentrate on interaction-critical regions while preserving human-scale structure elsewhere. Adaptive-HHI is the only method that satisfies all four criteria for dyadic human--humanoid interaction.

\begin{table}[th]
\centering
\footnotesize
\setlength{\tabcolsep}{4pt}
\renewcommand{\arraystretch}{1.2}
\caption{\textbf{Comparison of retargeting methods.} Adaptive-HHI extends OmniRetarget's interaction-mesh SOCP base with partner co-morphing (joint optimization over the partner SMPL-X) and distance-weighted, wrist-boosted adaptive Laplacian weighting for interaction and human-scale preservation.}
\label{tab:retarget-comparison}
\begin{tabularx}{\columnwidth}{@{}lCCCC@{}}
\toprule
\textbf{Method} & \makecell{Hard\\Kin.} & \makecell{Preserve\\Interact.} & \makecell{Partner\\Co-Morph} & \makecell{Adaptive\\Contact} \\
\midrule
IMMA \cite{imma2012}                                            & \cmark & \xmark     & \xmark     & \xmark     \\
PHC \cite{Luo2023PerpetualHC}                                   & \xmark     & \xmark     & \xmark     & \xmark     \\
GMR \cite{ze2025twistteleoperatedwholebodyimitation}            & \xmark     & \xmark     & \xmark     & \xmark     \\
VideoMimic \cite{allshire2025visualimitationenablescontextual}  & Soft       & \xmark     & \xmark     & \xmark     \\
OmniRetarget \cite{omniretarget2025}                            & \cmark & \xmark     & \xmark     & \xmark     \\
\rowcolor{gray!10}
\textbf{Adaptive-HHI Retargeting (Ours)}                                                   & \cmark & \cmark & \cmark & \cmark \\
\bottomrule
\end{tabularx}

\end{table}

\subsection{Pass~I Optimization}
\label{supp:passI}
Following the Laplacian-mesh formulation of \textsc{OmniRetarget}~\cite{ho2010spatial,omniretarget2025}, we build a per-frame Delaunay tetrahedralization over the ego's keypoints, matched semantically to humanoid keypoints, and couple the humanoid to the partner through contact-driving joints $\mathcal{P}=\{\text{L/R\_Wrist},\,\text{L/R\_Elbow}\}$. Let $\mathbf{q}_t \in \mathbb{R}^{36}$ be the humanoid configuration (29 DoF plus a 7D base pose), $\mathbf{x}_t(\mathbf{q}_t)$ and $\mathbf{x}^{\text{src}}_t$ the humanoid and ego mesh vertices, and $r_{j,t}(\mathbf{q}_t)$ and $d_{j,t}$ the distances from the humanoid's and the ego's joint $j\in\mathcal{P}$ to the partner's joint $j$. We solve, per frame:
\begin{align}
\min_{\mathbf{q}_t}\;
&\sum_{v} w_{v}\bigl\|\mathcal{L}(\mathbf{x}_t(\mathbf{q}_t))_v
  - s\,\mathcal{L}(\mathbf{x}^{\text{src}}_t)_v\bigr\|^{2}
+ \sum_{j\in\mathcal{P}} \lambda_j \bigl(r_{j,t}(\mathbf{q}_t) - d_{j,t}\bigr)^{2}
\nonumber\\
&+ \lambda_s \lVert\mathbf{q}_t-\mathbf{q}_{t-1}\rVert^{2}
+ \mathbf{q}_t^{\top}\mathbf{Q}\,\mathbf{q}_t,
\label{eq:passI}
\end{align}
subject to joint limits, a trust region on $\mathbf{q}_t$, and \textsc{OmniRetarget}'s velocity-based foot-sticking constraints. We optimize via SQP. The terms enforce \textbf{(i)}~\emph{local shape preservation}, matching the humanoid's Laplacian coordinates $\mathcal{L}(\cdot)_v$ at each keypoint $v$ to the ego's, scaled by the humanoid-to-ego height ratio $s$; \textbf{(ii)}~\emph{contact and reach preservation}, matching $r_{j,t}$ to $d_{j,t}$; \textbf{(iii)}~\emph{temporal smoothness} with respect to the previous solution, weighted by $\lambda_s$; and \textbf{(iv)}~\emph{posture regularization}, with diagonal $\mathbf{Q}\succeq 0$ (e.g.\ favoring neutral ankle pitch). The per-keypoint weights $w_v$ and per-joint weights $\lambda_j$ emphasize the wrists. Grounding is described in Sec.~\ref{supp:contact_aware_grounding}.

\subsection{Adaptive-HHI Retargeting Ablation}
Table~\ref{tab:retarget_ablation} ablates our retargeting on CHI3D, starting from OmniRetarget. Adding contact-adaptive weights reduces wrist error from 1.38 to 0.98\,cm and slightly raises contact preservation (0.915 to 0.933), but increases foot skating (0.009 to 0.014). Adding contact-and-reach preservation gives the best result on all three metrics: wrist error falls to 0.50\,cm, contact preservation reaches 0.936, and foot skating drops to 0.003. Overall, wrist error is reduced by about 64\% relative to the baseline, with most of the gain from the second component.
\begin{table}[h]
\centering
\caption{
\textbf{Adaptive retargeting ablation} on CHI3D. Mean $\pm$ std over clips. Best per column in bold.
}
\label{tab:retarget_ablation}
\footnotesize
\setlength{\tabcolsep}{2pt}
\resizebox{\linewidth}{!}{%
\begin{tabular}{lccc}
\toprule
& \textbf{Interaction}
& \textbf{Contact Preservation}
& \textbf{Foot Skating} \\
\cmidrule(lr){2-2}
\cmidrule(lr){3-3}
\cmidrule(lr){4-4}
\textbf{Method}
& $E_{\text{wrist}}$ [cm]~$\downarrow$
& Dur.~$\uparrow$
& Dur.~$\downarrow$ \\
\midrule
Baseline (OmniRetarget)
& 1.379\,$\pm$\,0.550 & 0.915\,$\pm$\,0.160 & 0.009\,$\pm$\,0.015 \\

+ contact-adaptive weights ($w_{v,t}$)
& 0.976\,$\pm$\,0.534 & 0.933\,$\pm$\,0.131 & 0.014\,$\pm$\,0.018 \\
\rowcolor{lightgray}
+ contact-and-reach preservation (Ours)
& \best{0.501\,$\pm$\,0.290} & \best{0.936\,$\pm$\,0.119} & \best{0.0034\,$\pm$\,0.010} \\
\bottomrule
\end{tabular}
}
\end{table}

\subsection{Contact-Aware Grounding}
\label{supp:contact_aware_grounding}
We observed that pinning the feet to the floor on every frame fails during dynamic maneuvers such as high kicks and vertical jumps, and that the Laplacian term does not constrain global vertical positioning. We resolve this by introducing contact-aware grounding via three complementary terms: (1)~\emph{velocity-based foot sticking}, inherited from \textsc{OmniRetarget}, holds a foot's horizontal position fixed only during stance intervals estimated from toe velocity; (2)~\emph{per-frame ground shifting} applies a smoothed correction to the robot base height, aligning with the floor during stance and tracking the source sole height during airborne phases, which we detect as runs of 3--30 frames with the sole more than 4\,cm above a rolling-minimum floor estimate; and (3)~\emph{ankle posture priors} combine foot-keypoint downweighting with a diagonal ankle-pitch prior to suppress toe-down artifacts arising from human--humanoid foot mismatches.

\subsection{Hyperparameters}
\label{app:hparams}
Table~\ref{tab:hparams} lists the hyperparameters used throughout the retargeting pipeline. All values were kept constant across all five source datasets. The distance weights reflect the contact semantics of dyadic interaction: wrist terms receive the largest weights since wrist placement dominates contact in our data (high-fives, handshakes, pushes), and elbow terms anchor the upper-arm orientation. The grounding parameters operate at the temporal scale of flight phases (3--30 frames at 30~Hz), separating genuine jumps from foot roll and capture drift, and the small smoothing width ($\sigma = 1$ frame) removes solver jitter without attenuating fast maneuvers such as kicks.

\begin{table}[t]
\centering
\small
\caption{\textbf{Retargeting hyperparameters.} Values are shared across all
datasets without per-clip tuning.}
\label{tab:hparams}
\begin{tabular}{lc}
\toprule
\textbf{Hyperparameter} & \textbf{Value} \\
\midrule
\multicolumn{2}{l}{\textit{\textbf{Pass I: Laplacian term}}} \\
Base keypoint weight & 10 \\
Wrist / foot keypoint multiplier & $\times 4$ / $\times 0.1$ \\
Laplacian target scale $s$ & $h_{\text{robot}}/h_{\text{ego}}$ \\
\midrule
\multicolumn{2}{l}{\textit{\textbf{Pass I: distance and regularization}}} \\
Wrist / elbow distance weight $\lambda_j$ & 50 / 25 \\
Smoothness weight $\lambda_s$ & 0.5 \\
Ankle-pitch prior & 2.0 \\
Shoulder-roll limit & $100^\circ$ \\
\midrule
\multicolumn{2}{l}{\textit{\textbf{Pass I: SQP}}} \\
Iterations per frame (first / subsequent) & 50 / 10 \\
Trust-region radius & 0.2 \\
Foot-sticking velocity threshold & 0.01~m/frame \\
\midrule
\multicolumn{2}{l}{\textit{\textbf{Grounding}}} \\
Rolling-minimum floor window & 61 frames \\
Airborne height threshold & 4~cm \\
Airborne run length & 3--30 frames \\
Ground-shift smoothing & $\sigma = 1$ frame \\
\midrule
\multicolumn{2}{l}{\textit{\textbf{Pass II: arm IK}}} \\
Contact threshold (source meshes) & 5~cm \\
Gauss--Newton iterations per frame & 10 \\
Target / source-pose / damping weight & 200 / 0.1 / 0.05 \\
Trust-region radius & 0.6~rad \\
\midrule
\multicolumn{2}{l}{\textit{\textbf{Frame rate}}} \\
Target rate & 30~Hz \\
\bottomrule
\end{tabular}
\end{table}

\section{Shared-Codebook VQVAE}

\subsection{Loss Ablation}

Table~\ref{tab:ablation-vqvae} ablates the two alignment losses used to build the shared human (H) and robot (R) codebook. Without alignment, the codebook reconstructs each embodiment well but translates poorly across them: H$\to$R and R$\to$H FD are 0.206 and 0.157, and no paired frames share a code. Adding both alignment losses reduces translation FD by more than an order of magnitude (to 0.0028 and 0.0044), lowers cross-embodiment reconstruction error (H$\to$R from 12.52 to 2.79\,cm and R$\to$H from 12.95 to 3.71\,cm), and raises code agreement from 0 to 0.78, while self-reconstruction slightly improves (H 3.67$\to$3.56\,cm, R 2.81$\to$2.61\,cm). Each loss alone helps (code agreement 0.51 continuous, 0.63 discrete), but the combination is best. Discrete alignment alone slightly hurts self-reconstruction (H 3.86\,cm, R 2.88\,cm) and H FD (0.0054), which the continuous loss recovers.

\begin{table}[t]
\centering
\footnotesize
\setlength{\tabcolsep}{3pt}
\renewcommand{\arraystretch}{1.1}
\caption{\textbf{Shared-codebook design ablation on HHI.} H and R denote human and robot motion; H$\to$R and R$\to$H denote encoding in one embodiment and decoding in the other. FD is Fréchet distance. MPJPE is the root-relative mean per-joint position error of the decoded pose against ground truth (cm; 21 SMPL-X body joints for H, 29 actuated joints of the G1 for R, hands excluded). Code Agree. is the fraction of paired H/R frames assigned the same code. Best per column in bold.}
\label{tab:ablation-vqvae}
\begin{tabular}{@{}lccccccccc@{}}
\toprule
& \multicolumn{4}{c}{\textbf{FD}~$\downarrow$} & \multicolumn{4}{c}{\textbf{MPJPE [cm]}~$\downarrow$} & \textbf{Code Agree.}~$\uparrow$ \\
\cmidrule(lr){2-5}\cmidrule(lr){6-9}\cmidrule(l){10-10}
\textbf{Method} & H & R & H$\to$R & R$\to$H & H & R & H$\to$R & R$\to$H & $H{=}R$ \\
\midrule
No alignment            & 0.0047 & 0.0026 & 0.2062 & 0.1570 & 3.665 & 2.809 & 12.520 & 12.951 & 0.000 \\
Continuous only         & \textbf{0.0045} & 0.0026 & 0.0034 & 0.0052 & 3.665 & 2.656 & 3.240 & 4.239 & 0.508 \\
Discrete only           & 0.0054 & 0.0031 & 0.0046 & 0.0054 & 3.855 & 2.876 & 3.063 & 4.036 & 0.634 \\
\midrule
\rowcolor{lightgray}
Both (Ours)             & 0.0047 & \textbf{0.0020} & \textbf{0.0028} & \textbf{0.0044} & \textbf{3.556} & \textbf{2.613} & \textbf{2.793} & \textbf{3.706} & \textbf{0.780} \\
\bottomrule
\end{tabular}
\end{table}

\subsection{Cross-Embodiment Translation: Human $\Rightarrow$ Robot}
\label{app:Cross-Embodiment Translation}

To translate a human motion into humanoid motion, we encode it with the human encoder $\mathrm{E}_h$, quantize it against the shared codebook $\mathcal{C}$, and decode the resulting tokens with the humanoid decoder $\mathrm{D}_r$. No humanoid motion is used at inference, and no optimization-based retargeting is run. Translation is therefore a single feed-forward pass through the tokenizer. We evaluate it on held-out paired clips from the HHI benchmark, where the retargeted humanoid motion serves as ground truth for the translated output.

\begin{figure}
    \centering
    \includegraphics[width=1\linewidth]{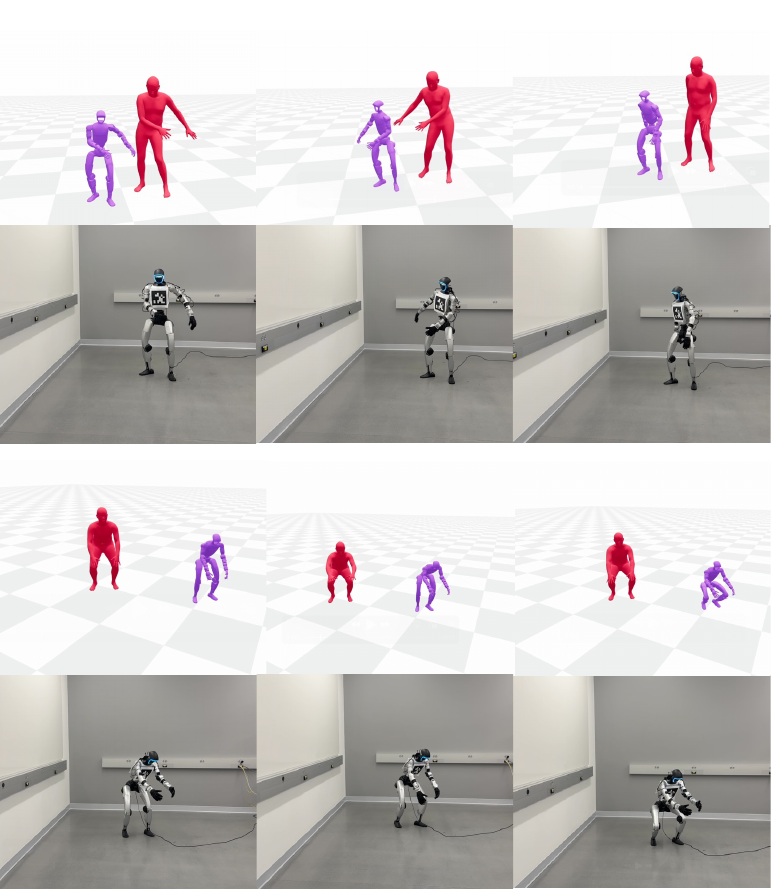}
    \caption{\textbf{Cross-embodiment motion translation with a shared-codebook VQ-VAE}. Human motion (red) is encoded into a discrete latent space shared between the human and humanoid embodiments, then decoded into humanoid motion (purple). Each example shows three frames of the input human motion alongside the translated humanoid motion in simulation (top row), and the same frames executed on the physical humanoid robot (bottom row).}
    \label{fig:Human_Robot}
\end{figure}

\noindent\textbf{Qualitative results}. Figure \ref{fig:Human_Robot} shows two translated sequences. The top rows of each example show the human input (red) and the translated humanoid motion (purple) in simulation. The bottom rows show the same motion tracked on the physical G1 by SONIC. In the first example (a standing, reaching gesture), the translated humanoid reproduces the arm extension and the timing of the reach. In the second example (a crouch), the humanoid matches the overall posture and descent to the floor. In both cases the output stays within the G1’s joint limits and proportions, so the motion is directly trackable on hardware. Because the human and humanoid have very different limb lengths and degrees of freedom, the translation does not copy joint angles one-to-one. It reproduces the pose through the humanoid’s own kinematics.

\noindent\textbf{Quantitative results}. Table \ref{tab:ablation-vqvae} reports translation quality alongside self-reconstruction. With both alignment losses, H→R translation reaches an FD of 0.0028 and a root-relative MPJPE of 2.79 cm. This is close to the humanoid’s own self-reconstruction error (2.61 cm) and is lower than the human’s self-reconstruction error (3.56 cm). Without alignment, the same shared codebook reconstructs each embodiment well (H: 3.67 cm, R: 2.81 cm) but translates poorly, with an H→R MPJPE of 12.52 cm and an FD of 0.206. This shows that sharing a codebook alone does not produce a transferable vocabulary, and that the alignment terms are what make the codes consistent across embodiments. Code agreement between paired human and humanoid frames rises from 0.00 to 0.78 with both losses.

\section{High-level Policy Details}
\label{app:high-level}

This section details the training procedures, hyperparameters, and network architectures for both the shared-codebook VQ-VAE and the DFoT components of our high-level policy.

\subsection{DFoT Training Loss}
We optimize \ourmethod{} across three complementary spaces---latent, motion, and reconstruction---to balance diffusion accuracy with kinematic fidelity:
\begin{equation}
\mathcal{L}_{\text{total}} = w_{\text{latent}} \mathcal{L}_{\text{latent}} + w_{\text{motion}} \mathcal{L}_{\text{motion}} + w_{\text{recon}} \mathcal{L}_{\text{recon}},
\end{equation}
with weighting coefficients $w_{\text{latent}}=1$, $w_{\text{motion}}=0.5$, and $w_{\text{recon}}=0.5$. The individual terms are

\begin{subequations}
\begin{align}
\mathcal{L}_{\text{latent}} &= \|\mathbf{z}^h - \hat{\mathbf{z}}^h\|_1 + \|\mathbf{z}^r - \hat{\mathbf{z}}^r\|_1, \\
\mathcal{L}_{\text{motion}} &= \sum_{a \in \{h, r\}} \| \mathbf{x}^a - \mathrm{D}_a(\hat{\mathbf{z}}^a) \|_1, \\
\mathcal{L}_{\text{recon}}  &= \sum_{a \in \{h, r\}} \| \mathbf{x}^a - \mathrm{D}_a(\mathrm{E}_a(\mathbf{x}^a)) \|_1.
\end{align}
\end{subequations}

\emph{Denoising Loss.} The primary supervision recovers the clean latents via a Smooth-$L_1$ loss between the target and predicted latents for both the human ($h$) and robot ($r$):
\begin{equation}
\mathcal{L}_{\text{latent}} = \|\mathbf{z}^h - \hat{\mathbf{z}}^h\|_1 + \|\mathbf{z}^r - \hat{\mathbf{z}}^r\|_1.
\end{equation}
 
\emph{Motion Loss.} To ensure physical plausibility, an auxiliary term decodes the predicted latents and applies a Smooth-$L_1$ loss in the raw motion space against the ground-truth motion features $\mathbf{x}^a$:
\begin{equation}
\mathcal{L}_{\text{motion}} = \sum_{a \in \{h, r\}} \| \mathbf{x}^a - \mathrm{D}_a(\hat{\mathbf{z}}^a) \|_1.
\end{equation}
 
\emph{Reconstruction Loss.} To enforce a robust latent bottleneck, we pass the ground-truth motion through a non-diffused autoencoding loop and penalize the round-trip reconstruction error:
\begin{equation}
\mathcal{L}_{\text{recon}} = \sum_{a \in \{h, r\}} \| \mathbf{x}^a - \mathrm{D}_a(\mathrm{E}_a(\mathbf{x}^a)) \|_1.
\end{equation}

\subsection{Implementation Details}
We train the shared-codebook VQ-VAE and DFoT models using the AdamW optimizer with an initial learning rate of $2\times10^{-4}$, a weight decay of $1\times10^{-4}$, and a mini-batch size of 256. The learning rate is cosine-decayed to zero over the course of training. Both the Shared-Codebook VQ-VAE and DFoT architectures utilize GELU activations and LayerNorm layers in their feed-forward blocks, and the DFoT model relies on a standard post-norm Transformer encoder architecture. 

For the DFoT, we adopt a discrete diffusion formulation with an $x_0$-prediction parameterization. During training, we sample a diffusion timestep $t \in \{1, \dots, 1000\}$ independently for each agent token to prevent architectural shortcutting, training the network to reconstruct the corresponding clean motion token $x_0$. At inference time, we employ deterministic DDIM sampling over 30 steps. Unless otherwise stated, each VQ-VAE model is trained for 100,000 steps, and each DFoT model is trained for 300,000 steps. Final checkpoints are selected based on the lowest validation loss.

All training experiments are conducted on a single NVIDIA RTX A6000 GPU with 48\,GB of VRAM. Under the configuration detailed in Table~\ref{tab:net_details}, a full DFoT model convergence requires approximately 1 day of training time.

\subsection{Network Configurations}

Table~\ref{tab:net_details} summarizes the hyperparameters of both sub-modules, shared-codebook VQ-VAE and DFot.

\noindent\textbf{Shared-codebook VQ-VAE.}
Each agent's motion is encoded by an embodiment-specific 1D convolutional encoder operating along the temporal axis with a compression factor of $\omega{=}4$, mapping per-agent pose features of dimension $D_\text{in}$ (with $D_h$ for the human and $D_r$ for the humanoid) into a shared latent space of dimension $d_\text{vq}{=}512$. Both embodiments quantize against a single codebook of $K{=}512$ entries updated via exponential moving average, which encourages a common motion vocabulary across the human--humanoid embodiment gap; embodiment-specific decoders mirror their encoders to reconstruct each agent's pose representation.

\noindent\textbf{DFoT.}
The high-level policy operates on unified motion tokens of dimension $D_\text{m}$, which augment the VQ latents with localized cross-agent transformations ($4 \times 9 \times P$ pairwise relative features). Per-token noise levels and agent identity are injected through sinusoidal and learned embeddings, respectively, concatenated with the motion tokens and projected to the model dimension $d{=}512$. Temporal position is encoded with RoPE, and the core consists of 6 Transformer blocks with 8 attention heads attending jointly over time and agents (sequence length $T' \times P$). A linear head maps back to $D_\text{m}$.

\begin{table}[t]
  \centering
  \footnotesize
  \setlength{\tabcolsep}{3pt}
  \renewcommand{\arraystretch}{1.15}
  \caption{\textbf{Network configuration details.} $T$ is the raw sequence length and $T' = T/\omega$ the number of tokens after temporal compression ($\omega{=}4$); $P$ is the number of interacting agents; $D_\text{in}$ is the per-agent pose dimension ($D_h$ or $D_r$); $D_\text{c}$ is the local conditioning dimension; $d_\text{e}$ is the embedding dimension; and $D_\text{m} = d_\text{vq} + 4 \times 9 \times P$ is the unified motion token dimension tracking localized cross-agent transformations.}
  \label{tab:net_details}
  \begin{tabularx}{\columnwidth}{@{}
      >{\raggedright\arraybackslash}p{0.13\columnwidth}
      >{\raggedright\arraybackslash}p{0.16\columnwidth}
      l
      >{\raggedright\arraybackslash}X@{}}
    \toprule
    \textbf{Module} & \textbf{Component} & \textbf{Input shape} & \textbf{Operation \& hyperparameters} \\
    \midrule
    \multirow{3}{=}{\textbf{Shared-Codebook VQ-VAE}}
      & Human \& humanoid encoder
      & $(T, P, D_\text{in}{+}D_\text{c})$
      & 1D Conv + ResNet, 2 layers, $d_\text{vq}{=}512$ \\
      & Codebook
      & $(T', P, d_\text{vq})$
      & VQ, $K{=}512$ codes, EMA updates \\
      & Human \& humanoid decoder
      & $(T', P, d_\text{vq}{+}D_\text{c})$
      & Mirror of encoder, 2 layers \\
    \midrule
    \multirow{6}{=}{\textbf{DFoT}}
      & Noise emb.
      & $(T', P, d_\text{e})$
      & Sinusoidal, $d_\text{e}{=}256$ \\
      & Agent emb.
      & $(T', P, d_\text{e})$
      & Learned, $d_\text{e}{=}256$ \\
      & Input proj.
      & $(T', P, D_\text{m}{+}2d_\text{e})$
      & Linear + LayerNorm, 3 layers, ${\rightarrow}\,d{=}512$ \\
      & Time emb.
      & $(T', P, d)$
      & RoPE on hidden dim $d$ \\
      & Core blocks
      & $(T'{\times}P, d)$
      & Transformer, 6 layers, 8 heads, MLP $4d$ \\
      & Output head
      & $(T', P, d)$
      & Linear, $d \rightarrow D_\text{m}$ \\
    \bottomrule
  \end{tabularx}
\end{table}

\section{End-to-end System Details}
\label{app:system}

We implement a real-time system for Social Humanoid interaction. The system consists of a unified robot-human perception stack, followed by the high-level and low-level policies. This section describes the perception stack, system architecture, and the end-to-end latency.

\subsection{Camera Setup and Calibration}

We use an exoview ZED 2i camera to capture RGB+D at 1280x720 and 15~Hz (the capture rate is bandwidth-limited, see Sec. ~\ref{app:system:arch}). To estimate the gravity direction $\mathbf{n}$ and camera-to-floor-distance $d$, we capture 30 depth frames, apply per-pixel temporal median, and backproject to a point cloud in the camera optical frame. We fit a plane $\mathbf{n}^\top \mathbf{p} = d$ with RANSAC under the constraints $\|\mathbf{n}\|=1$ and $d>0$, and refine the winning hypothesis via SVD on the inliers. To separate the ground plane from other flat surfaces, we: (i) enforce gravity direction $\mathbf{n}^\top \bar{\mathbf{p}} > 0$ for the inlier centroid $\bar{\mathbf{p}}$, (ii) use directionality heuristic requiring a normal that is sufficiently vertical, and (iii) require a minimum inlier count.

\begin{wraptable}[19]{R}{0.48\linewidth}
\centering
\small
\setlength{\tabcolsep}{8pt}
\vspace{-12px}
\caption{\textbf{End-to-End Latency.}}
\begin{tabular}{lrr}
\toprule
\textbf{Stage} & \textbf{p50} (ms) & \textbf{p95} (ms) \\
\midrule
\textbf{perception}  & 104.4 & 120.4 \\
\quad fetch          & 7.9   & 8.4   \\
\quad sync           & 20.3  & 21.2  \\
\quad wire           & 52.7  & 66.4  \\
\quad JPEG+zstd      & 23.5  & 24.4  \\
\textbf{human}       & 79.4  & 115.6  \\
\textbf{high-level}  & 103.2 & 118.2 \\
\quad interpolate    & 38.6  & 57.8  \\
\quad VQ-VAE         & 8.1   & 10.4  \\
\quad DFoT           & 50.4  & 58.5  \\
\rowcolor{gray!30}
\quad postproc       & +133  & ---   \\
\textbf{low-level}   & 21.1  & 22.0  \\
\quad SONIC          & 0.9   & 1.0   \\
\quad wire           & 20.2  & 21.0  \\
\rowcolor{gray!30}
\quad wire buffer    & +100  & ---   \\
\midrule
\textbf{total} (min) & 308.1 & 376.2 \\
\textbf{total} (max) & 541.1 & 609.2 \\
\bottomrule
\end{tabular}
\label{tab:perception-latency}
\end{wraptable}

\subsection{Global Human Estimation}

Given per-frame RGB+D, we want to estimate the human's SMPL pose and full SE(3) pelvis pose in a gravity-aligned, floor-origin camera frame $C_\text{gcam}$. We detect humans with YOLOv11 and pass each detection to Fast Sam 3D Body (FS3DB) \cite{yang2026fastsam3dbody}, which consists of an optimized MHR head and an MHR-to-SMPL regressor. Because the regressor is not trained to preserve global orientation, we recover orientation by using the Kabsch algorithm between corresponding regressed SMPL vertices and MHR vertices in the camera frame. To recover translation, we query a depth patch at selected vertices on the torso to get estimates for the pelvis depth. We perform outlier rejection using $z$-buffer occlusion and MAD, aggregate the predictions by the detector's confidence on the vertices, and then backproject the 2D pelvis using the camera intrinsics. A constant-velocity Kalman filter smooths the resulting trajectory against noisy depth. This yields $T_{h \to C_\text{cam}}$ for every detected human, which we convert to $T_{h \to C_\text{gcam}}$ using $(\mathbf{n}, d)$.

To maintain identities across frames, we construct and run a Kalman filter per tracked human. For each incoming frame, we KF-predict each active track and use the Hungarian algorithm on 3D Euclidian distance to assign each detected human to a track. Unassigned humans are allocated new tracks, and unassigned tracks retire after several consecutive misses. The robot maintains interaction with a single track determined on startup by the closest human. If no humans are visible, the high-level policy can switch from partner-conditioned prediction to joint prediction.

\subsection{Global Robot Estimation}

We place 8x8 inch AprilTags from the 36h11 family on the front and rear of the robot. During an initialization phase, we use the AprilTags to derive an estimate of the robot's SE(3) pelvis pose $T_{r_0 \to C_\text{cam}}$ and similarly use $(\mathbf{n}, d)$ to convert to $T_{r_0 \to C_\text{gcam}}$. The robot is positioned so that at least one tag is unoccluded during calibration. At inference time, we use an onboard, proprioception-based state estimation available in the Unitree SDK. Because the estimator is with respect to some canonical frame determined on robot startup, we use the estimator to compute $T_{r \to r_0}$, and compose it with the calibration estimate to get $T_{r \to C_\text{gcam}}$. Empirically, the Unitree estimator is sufficiently stable for our interaction scenarios.

\subsection{System Architecture}
\label{app:system:arch}

A laptop connected to the ZED 2i and the Unitree G1 performs time synchronization and publishes the RGB+D images and the robot estimation plus proprioception as a bundle at 15~Hz. Due to FS3DB's compute requirements, we connect the laptop to a remote server using WiFi. We JPEG-compress the RGB stream and apply Zstd compression on the depth stream to maximize streaming rate. We run two instances of FS3DB with a global estimation coordinator at 15~Hz to tolerate variations in GPU load. The coordinator bundles the human and robot estimates and sends them to the high-level policy, which interpolates the incoming data to 30~Hz, performs inference, and emits chunks of 30~Hz robot motions at 7.5~Hz. Finally, SONIC interpolates the motions to 50~Hz and tracks them by providing desired DoF angles for a PD controller. We run inference on 2x NVIDIA 4090 GPUs for global human estimation, 1x NVIDIA 4090 GPU for the high-level policy, and 1x NVIDIA 4060 Mobile GPU for the low-level policy.

We support two optional improvements upon the existing system which tradeoff latency to improve motion generation quality and stability. After DFoT inference, we delay the emission of robot motion chunks by 1 chunk to improve decoded motion continuity. For robustness against deviations in network latency and other sources of variation, SONIC maintains a buffer of previous motion and delays execution by 100~ms.

\subsection{Latency Experiments}

We evaluate the end-to-end latency of our pipeline. In Table \ref{tab:perception-latency}, we report the 50 and 95-percentile latencies ($n = 200$), splitting the evaluation by stage. Stage 1: \textbf{perception}, which includes (i) \textit{fetch}: retrieving the RGB+D image from the camera, (ii) \textit{sync}: synchronization with robot estimation and construction of bundle, (iii) \textit{wire}: network and blocking on input on the server-side, (iv) \textit{JPEG+zstd}: compression encoding and decoding. Stage 2: \textbf{human} estimation, which bundles distributed coordination, FS3DB inference, and global estimation. Stage 3: \textbf{high-level} policy, which includes (i) \textit{interpolate}: interpolation of SMPL from 15$\to$30~Hz, (ii) \textit{VQ-VAE}: encoding and decoding, (iii) \textit{DFoT}: DFoT inference, (iv) \textit{postproc} (optional): buffered decoding to improve motion continuity. Stage 4: \textbf{low-level} policy, which includes (i) \textit{SONIC}: construction of observations, policy inference, sending commands to G1, (ii) \textit{wire}: network transport back to laptop, (iii) \textit{wire buffer} (optional): execution delay to improve tolerance to variation. We highlight optional components which tradeoff motion generation quality for latency. In reactive regimes, we achieve a median latency of 308ms.

\subsection{System Safety} 
We validate safety progressively, from simulation to a gantry-supported robot to gantryless deployment. SONIC holds a standing pose until a dedicated operator, separate from the person interacting with the robot, engages it, and then smoothly executes high-level policy (HLP) frames. We clamp the desired yaw change to $45^\circ$/s. SONIC smoothly returns to standing if consecutive HLP predictions differ by more than $0.7\,\text{rad}$, or if the operator judges the behavior unsafe. Across five 60\,s rollouts, we observed no unsafe generated behavior. Note that the low-level policy is neither force-adaptive nor force-compliant, so physical contact is not actively regulated.